%% file: main.tex
\documentclass[10pt,twocolumn,letterpaper]{article}

\usepackage{wacv}
\usepackage{times}
\usepackage{epsfig}
\usepackage{graphicx}
\usepackage{amsmath}
\usepackage{amssymb}
\usepackage{accsupp}
\usepackage[accsupp]{axessibility}  

\usepackage{comment}
\usepackage{enumitem}
\usepackage[normalem]{ulem}
\useunder{\uline}{\ul}{}
\newcommand\Tstrut{\rule{0pt}{2.2ex}}         
\usepackage{booktabs}
\usepackage{bbm}
\newcommand{\Lagr}{\mathcal{L}}

\makeatletter
\@namedef{ver@everyshi.sty}{}
\makeatother
\usepackage{tikz}
\usepackage{pgfplots}
\pgfplotsset{compat=newest}
\usetikzlibrary{backgrounds}
\pgfdeclarelayer{foreground}
\pgfsetlayers{main,foreground}

\usepackage{xcolor}
\definecolor{correctgreen}{RGB}{53, 140, 30}
\definecolor{wrongred}{RGB}{201, 43, 43}
\definecolor{rgbblue}{RGB}{165, 200, 255}
\definecolor{floworange}{RGB}{255, 180, 109}
\definecolor{codegreen}{RGB}{118,181,197}
\definecolor{codedarkergreen}{RGB}{4, 99, 110}
\definecolor{codepurple}{rgb}{0.58,0,0.82}
\definecolor{backcolour}{rgb}{0.95,0.95,0.92}
\definecolor{commentgreen}{RGB}{2,112,10}
\definecolor{eminence}{RGB}{108,48,130}
\definecolor{weborange}{RGB}{163, 113, 55}
\definecolor{frenchplum}{RGB}{129,20,83}

\usepackage{listings}
\lstdefinestyle{mystyle}{breaklines=true,
  basicstyle=\color{black}\scriptsize\ttfamily, 
  linewidth=\columnwidth,
  columns  = fullflexible,
  frame = tb,
  commentstyle=\color{codedarkergreen},
  morekeywords={softmax, sinkhorn},
  classoffset=1,
  keywordstyle=\color{magenta},
  language=python}
\usepackage{fnpct}
\makeatletter 
\renewcommand{\@makefntext}[1]{%
  \setlength{\parindent}{4pt}%
  \begin{list}{}{\setlength{\labelwidth}{2.8mm}
    \setlength{\leftmargin}{\labelwidth}%
    \setlength{\labelsep}{0.4pt}%
    \setlength{\itemsep}{0pt}%
    \setlength{\parsep}{0pt}%
    \setlength{\topsep}{0pt}%
    \footnotesize}%
  \item[\@makefnmark]#1
  \end{list}%
}
\makeatother 

\makeatletter 
\renewcommand\maketitle{\par
  \begingroup
    \renewcommand\thefootnote{\@fnsymbol\c@footnote}%
    \def\@makefnmark{\rlap{\@textsuperscript{\normalfont\@thefnmark}}}%
    \long\def\@makefntext##1{\parindent -5em \noindent
            \hb@xt@1.0em{%
                \hss\@textsuperscript{\normalfont\@thefnmark \nobreak\hspace{0.4pt}}}##1}%
    \if@twocolumn
      \ifnum \col@number=\@ne
        \@maketitle
      \else
        \twocolumn[\@maketitle]%
      \fi
    \else
      \newpage
      \global\@topnum\z@   
      \@maketitle
    \fi
    \thispagestyle{plain}\@thanks
  \endgroup
  \setcounter{footnote}{0}%
  \global\let\thanks\relax
  \global\let\maketitle\relax
  \global\let\@maketitle\relax
  \global\let\@thanks\@empty
  \global\let\@author\@empty
  \global\let\@date\@empty
  \global\let\@title\@empty
  \global\let\title\relax
  \global\let\author\relax
  \global\let\date\relax
  \global\let\and\relax
}
\makeatother 

 \def\wacvPaperID{59} 

\wacvfinalcopy 

\ifwacvfinal
\def\assignedStartPage{1} 
\fi

\ifwacvfinal
\usepackage[breaklinks=true,bookmarks=false]{hyperref}
\else
\usepackage[pagebackref=true,breaklinks=true,colorlinks,bookmarks=false]{hyperref}
\fi

\ifwacvfinal
\else
\fi

\begin{document}
\title{Self-supervised Video Representation Learning \\with Cross-Stream Prototypical Contrasting}

\author{Martine Toering\! $^1$\thanks{Correspondence to: \scriptsize{\tt{\href{mailto:martine.toering@gmail.com}{martine.toering@gmail.com}}}.} \ \ \ \ Ioannis Gatopoulos\! $^2$ \ \ \ \ Maarten Stol\! $^2$ \ \ \ \ Vincent Tao Hu\! $^1$ \vspace{0.4em}\\
\normalsize{$^1$\! University of Amsterdam \ \ \ \ \ \ $^2$\! BrainCreators B.\!V.} \\
}


\maketitle

\input{sections/_abstract}
\input{sections/0-introduction}

\input{sections/1-related_works}

\input{sections/2_method}
\input{sections/3_experiment}
\input{sections/4_conclusion}

\clearpage
{\small
\bibliographystyle{ieee_fullname}
\bibliography{egbib}
}

\clearpage

\begin{center}
\subsection*{Supplementary Material for \\ Self-supervised Video Representation Learning \\with Cross-Stream Prototypical Contrasting \\ }    
\end{center}

\appendix
\input{sections/appendix}

\end{document}

%% file: sections/_abstract.tex
\begin{abstract}
Instance-level contrastive learning techniques, which rely on data augmentation and a contrastive loss function, have found great success in the domain of visual representation learning. They are not suitable for exploiting the rich dynamical structure of video however, as operations are done on many augmented instances. In this paper we propose ``Video Cross-Stream Prototypical Contrasting", a novel method which predicts consistent prototype assignments from both RGB and optical flow views, operating on sets of samples. Specifically, we alternate the optimization process; while optimizing one of the streams, all views are mapped to one set of stream prototype vectors. Each of the assignments is predicted with all views except the one matching the prediction, pushing representations closer to their assigned prototypes. As a result, more efficient video embeddings with ingrained motion information are learned, without the explicit need for optical flow computation during inference. We obtain state-of-the-art results on nearest-neighbour video retrieval and action recognition, outperforming previous best by +3.2\% on UCF101 using the S3D backbone (90.5\% Top-1 acc), and by +7.2\% on UCF101 and +15.1\% on HMDB51 using the R(2+1)D backbone.\footnote{\noindent Code is available at \scriptsize\url{https://github.com/martinetoering/ViCC}.}
\end{abstract}

%% file: sections/0-introduction.tex
\section{Introduction}
\begin{figure}[t]
\begin{center}
   \includegraphics[width=0.9\columnwidth]{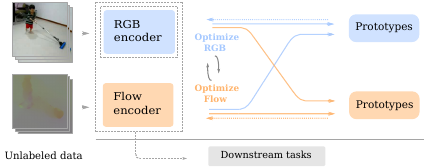}
\end{center}
   \caption{\textbf{RGB and optical flow} are used as two streams in the training of one stream by predicting consistent prototype assignments from features. By also alternating the training, we transfer knowledge cross-stream from motion (flow) to appearance (RGB) useful for downstream video tasks with optional optical flow.} 
   \label{fig:method_small}
\end{figure}

The goal of this paper is self-supervised representation learning for video. Visual representation learning methods based on instance-level contrasting have significantly reduced the gap with supervised learning in image-based tasks \cite{chen_simple_2020, he_momentum_2020, oord_representation_2019} and video \cite{qian_spatiotemporal_2021,  han_self-supervised_2020}. These contrastive learning frameworks require an augmentation module that obtains multiple views of one instance, and a loss function that contrasts between augmented views of instances. The objective can be viewed as instance discrimination: producing higher similarity scores between augmentations of the same instances, rather than with those that belong to different ones (\textit{negative examples}). As a result, the methods rely heavily on data augmentation in order to learn powerful representations. Furthermore, a vast amount of negative examples has to be obtained which often relies on either memory banks \cite{he_momentum_2020} or large batch sizes \cite{chen_simple_2020}. 

To adopt these techniques into the video domain efficiently, we make the following observations. First, we notice that though video also provides natural augmentation with viewpoint changes, illumination (jittering) and deformation, still spatiotemporal coherence and motion are not explicitly used. We are inspired by the two-streams hypothesis for vision processing in the brain \cite{schneider_two_1969, GoodaleMelvynA1992Svpf}, suggesting two pathways: the ventral stream involved in object recognition and the dorsal stream locating objects and recognizing motion. Motion without appearance information can be a rich source of information for humans \cite{JohanssonGunnar1973Vpob}, however more recent works propose that it is likely that the streams interact \cite{McIntoshRobertD2009Tvsf}. Second, we believe instance-level contrastive learning is inefficient and neglects the use of semantic similarity between instances. Low similarity scores are produced for a large pool of negative pairs regardless of their semantic similarity, resulting in undesirable distances between samples in the embedding. To resolve this, several works have explored alternatives to random sampling for negative examples \cite{cao_parametric_2020, chuang_debiased_2020}, such as hard negative mining \cite{kalantidis_hard_2020-1, robinson_contrastive_2021}. We are instead interested in leaving instance-level comparisons and include mappings to \textit{prototypes} (defined as representatives of semantically similar groups of features), providing a possible benefit on video representations without any potential costs from distance searches in the data.

In this work, we present a novel self-supervised method called \noindent \textit{\noindent \textbf{Vi}deo \noindent \textbf{C}ross-Stream Prototypical \noindent \textbf{C}ontrasting (ViCC)} where we consider RGB and optical flow as distinct views for video contrastive learning, to influence appearance and motion learning respectively. The two input streams and spatiotemporal augmentations are united into one framework. In each iteration of the optimization of one stream, views are assigned to a set of prototypes and assignments are subsequently predicted from the features, see Figure \ref{fig:method_small}. 

Our contributions can be summarized as below.
\begin{itemize}[topsep=0pt,itemsep=-1ex,partopsep=1ex,parsep=1ex, leftmargin=*]
\item We introduce a novel visual-only self-supervised learning framework for video that contrasts using sets of views from two streams (RGB and flow). We demonstrate the benefits of operating on stream prototypes over contrastive instance learning, avoiding unnecessary comparisons and hence computations, while improving accuracy. 
\item We propose a new training mechanism for video, in which RGB and flow streams are interconnected in two ways: prototypes are predicted from both streams and the optimization process is alternated. As motion information is transferred to the RGB model, we can discard the optical flow network in deployment scenarios depending on speed and efficiency requirements. 
\item We perform extensive ablation studies to provide an in-depth analysis of our method. Our result reaches state-of-the-art on UCF101 \cite{soomro_ucf101_2012} and HMDB51 \cite{kuehne_hmdb_2011} on the two backbones S3D \cite{xie_rethinking_2018} and R(2+1)D \cite{tran_closer_2018}.
\end{itemize}

%% file: sections/1-related_works.tex
\section{Related work}
\textbf{Contrastive instance learning.} Instance discrimination considers each sample as its own class in the data. As such a classifier becomes computationally infeasible fast, \cite{wu_unsupervised_2018} use noise-contrastive estimation \cite{gutmann_noise-contrastive_2012} and a memory bank to store representations as their pool of negative samples. Other solutions include the work from Chen \etal \cite{chen_improved_2020}, which retrieves more negative samples by using large batch sizes. He \etal \cite{he_momentum_2020} propose a momentum encoder with a dynamic dictionary look-up. Another line of work contrasts between the global image and local patches \cite{oord_representation_2019, hjelm_learning_2019}. Our method instead uses complementary modalities as main views and intuitively learns its own positive and negative examples from both feature spaces through the prototypes. Contrasting is done between instances and prototypes, going beyond instance-level learning while avoiding the need for substantial batch sizes \cite{chen_simple_2020} or large memory banks \cite{he_momentum_2020}. 

\textbf{Clustering in latent space.} Combining clustering with representation learning to obtain pseudo-labels has been proposed in various self-supervised learning settings \cite{yan_clusterfit_2020, caron_deep_2018, asano_labelling_2020, caron_unsupervised_2020, li_prototypical_2021}. Asano \etal \cite{asano_self-labelling_2020} propose a solution of degenerate solutions by casting clustering into an instance of the optimal transport problem. Caron \etal \cite{caron_unsupervised_2020} use this clustering setup in a contrastive learning setting by enforcing consistency between different views, comparing cluster assignments instead of individual features. Furthermore, an online clustering and simultaneous feature learning mechanism was proposed in \cite{zhan_online_2020}. Our objective is most similar to \cite{caron_unsupervised_2020} and \cite{li_prototypical_2021}, aligning cluster assignments for augmented instances in an online manner. However, we apply our method on video, use augmentation in the form of optical flow and alternate the training of models and prototypes to incorporate information in both streams.

\textbf{Self-supervised video and distillation.} Advances in 3DConvNets \cite{tran_learning_2015, hara_can_2018, tran_closer_2018} have driven video research forwards. Self-supervised approaches exploring pretext tasks are often based on the temporal domain, such as the order of frames or clips \cite{misra_shuffle_2016, fernando_self-supervised_2017, lee_unsupervised_2017, xu_self-supervised_2019}, learning the arrow of time \cite{pickup_seeing_2014, wei_learning_2018} or pace \cite{yao_video_2020,  cho_self-supervised_2020, wang_self-supervised_2020, benaim_speednet_2020}. Pretext tasks that were previously explored in the image domain have been proposed and extended \cite{jing_self-supervised_2019, kim_self-supervised_2019}. Other approaches include leveraging the consistency in frames by temporal correspondence \cite{lai_mast_2020, lai_self-supervised_2019}, tracking patches \cite{wang_unsupervised_2015, wang_learning_2019}, future frame prediction \cite{goroshin_unsupervised_2015, vondrick_generating_2016} or future feature prediction \cite{han_video_2019, han_memory-augmented_2020}. Multiple works explore optical flow for self-supervision \cite{han_memory-augmented_2020, han_self-supervised_2020, mahendran_cross_2018}. A cross-stream approach was first proposed by \cite{ohnishi_improved_2016}. As opposed to them, we use contrastive learning without dense trajectories. Mahendran \etal \cite{mahendran_cross_2018} use optical flow as supervision for RGB. Tian \etal \cite{tian_contrastive_2020} first explore the use of RGB and optical flow as views for contrastive learning. Most similar to our work are \cite{tian_contrastive_2020} and \cite{han_self-supervised_2020}, which both employ RGB and flow in a two-stream manner for contrastive learning. Han \etal \cite{han_self-supervised_2020} use an alternated training process and samples hard positive examples from the other stream. Different from these works, we do not employ instance-level contrastive learning. As we use prototype mappings of our features and subsequently predict feature assignments, our streams leverage a stronger interplay. We also incorporate informed negative examples from both streams through our prototypes and we do not use a momentum encoder \cite{he_momentum_2020}. As optical flow computation can be computationally expensive, several works avoid flow computation during inference while utilizing it during training, \eg through knowledge distillation \cite{stroud_d3d_2020, crasto_mars_2019, zhao_dance_2019, gavrilyuk_motion-augmented_2021} which is related to our work. Our proposed method instead keeps two streams and leverages an alternated optimization process to perform a form of distillation through contrastive learning, avoiding the need for optical flow while still enabling its optional use. 

\textbf{Multi-modal approaches.} Video allows for a multi-modal approach by using information such as audio \cite{alwassel_self-supervised_2020, asano_labelling_2020} and text \cite{miech_end--end_2020, sun_learning_2019} to learn from correspondence between modalities. Alwassel \etal \cite{alwassel_self-supervised_2020} use a cross-modal audio-video iterative clustering and relabeling algorithm. Asano \etal \cite{asano_labelling_2020} employ both RGB and audio in a simultaneous clustering and representation learning setting, following \cite{asano_self-labelling_2020}. Our method strictly speaking does not leverage multiple modalities as we use an optical flow representation originally extracted from the RGB representation, without introducing any external information. However, our work similarly leverages the interplay of complementary information and could therefore be used alternatively as a multi-modal approach, \eg leveraging audio in addition to optical flow in order to improve representations further.

%% file: sections/2_method.tex
\section{Method}
\label{sec:method}

\begin{figure*}[ht]
\begin{center}
    \resizebox{0.85\textwidth}{!}{%
   \includegraphics[width=1\linewidth]{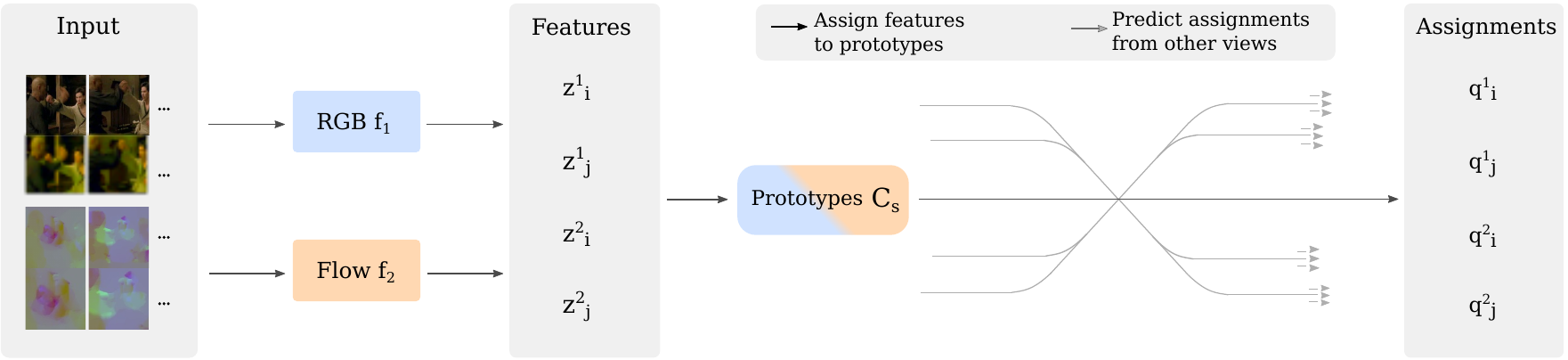}
   }
\end{center}
   \caption{\textbf{Video Cross-Stream Prototypical Contrasting.} Two different augmented samples are obtained for both RGB and flow. The encoders $f_1$ and $f_2$ map samples from RGB and flow respectively to obtain features $z^{1}_{i}, z^{1}_{j}, z^{2}_{i}, z^{2}_{j}$, which are in turn assigned to either RGB or flow prototype vectors, depending on which stream $s \in \{1,2\}$ is optimized. Next, the stream prototype assignments $q^{1}_{i}, q^{1}_{j}, q^{2}_{i}, q^{2}_{j}$ are predicted using features only from the three other views. The encoder and prototypes from the optimized stream are updated by backpropagation, while the other encoder remains fixed.}
\label{fig:method}
\end{figure*}
We first introduce preliminaries on instance-level contrastive learning in Section~\ref{sec:preliminaries}. We explain how we can use RGB and optical flow separately to predict and learn prototypes following \cite{caron_unsupervised_2020} in Section~\ref{sec:predict}. Finally, we introduce our contribution which consists of the cross-stream interplay and the steps of our algorithm in Section \ref{sec:crossstream}.

\subsection{Preliminaries}
\label{sec:preliminaries} 
Contrastive instance learning \cite{he_momentum_2020, chen_simple_2020} can be defined as a self-supervised learning method which contrasts in the latent space by maximizing agreement between different augmented views of the same data instances. Three key components in this framework are \textit{i)} a data augmentation module that transforms a given sample $x$ into two views $x_i$ and $x_j$ by applying separate transformations $t$ and $t'$ sampled from the set of augmentations $T$, \textit{ii)} the embedding function $f(\cdot)$ consisting of an encoder and a small MLP projection head that extracts feature vectors $z_i$ and $z_j$ from views, and \textit{iii)} a contrastive loss function that contrasts between $x_i$ and a set $\{x_k\}$ of augmented pairs that includes our positive pair. Given a dataset $X = \{x_1, x_2, ..., x_n\}$, we aim to learn a function $f(\cdot)$ that maps X to $Z = \{z_1, z_2, ..., z_n\}.$ The contrastive loss objective for a positive pair $(i, j)$, referred to as the InfoNCE loss \cite{sohn_improved_2016, oord_representation_2019, chen_simple_2020}, is then given by 
\begin{equation}
    \Lagr^{\text{InfoNCE\ }} (z_i, z_j) = - \ \text{log} \ \dfrac{\text{exp}(z_i \cdot z_j / \tau )}{\sum_{k \neq i} \text{exp}(z_i \cdot z_k / \tau)},
\end{equation}
where $\tau$ is the temperature hyperparameter and $z_i$ $\cdot$ $z_j$ refers to the dot product between normalized vectors, \ie cosine similarity. The final loss is computed for all available positive pairs. Given a positive pair, a sufficiently large number of negative examples in $\{x_{k}\}$ needs to be available for which storage of features besides the mini-batch is often needed. The contrastive learning mechanism also neglects to take into account the informativeness of samples.

\subsection{Predicting stream prototype assignments}
\label{sec:predict}
In our proposed method we avoid instance-level contrasting by using for each stream a set of \textit{prototypes} in our contrasting. Furthermore, we extend the augmentation module by considering RGB frames and optical flow as views. Mathematically, given a video clip $x$ we first consider the two streams as views, obtaining ${x} = \{x^{1}, x^{2}\}$ which describe a RGB and a flow sample respectively. The objective is to learn the stream representations $z^{1} = f_{1}(x^{1}$) and $z^{2}=f_{2}(x^{2})$ through learning their encodings $f_1(\cdot)$ and $f_2(\cdot)$. Each of the encoders has a set of $K$ trainable prototype vectors, $\{c^{1}_1, ..., c^{1}_K\}$ $\in$ $C_{1}$ and $\{c^{2}_1, ..., c^{2}_K\}$ $\in$ $C_{2}$, implemented as a linear layer in the networks. 

Consider only the training of one encoder $f_s$ on its own stream $s$ where $s \in \{1,2\}$. We denote the corresponding prototype set as matrix $C_{s}$ with columns $c^{s}_1, ..., c^{s}_k$. Given input sample $x^s$, we obtain two augmented versions $\{x^{s}_{i}, x^{s}_{j}\}$. After applying the encoder $f_{s}(\cdot)$ we obtain features $\{z^s_{i}, z^{s}_{j}\}$. The features are mapped to the set of prototypes $C_{s}$ to obtain cluster assignments $\{q^{s}_{i}, q^{s}_{j}\}$, as detailed in the following section. The features and assignments are subsequently used in the following prediction loss: 
\begin{equation}
\begin{gathered}
\label{eq:single}
    \Lagr^{\text{Single-stream}}_{\text{s}}(z^{s}_{i}, z^{s}_{j}) = l_s\big(z^{s}_{j}, q^s_{i}\big) + l_s\big(z^s_{i}, q^{s}_{j}\big) \ . 
\end{gathered}
\end{equation}
Each of the terms represents the cross-entropy loss between the stream prototype assignment $q$ and the probability obtained by a softmax on the similarity between $z$ and $C_s$: 
\begin{equation} 
\begin{gathered}
    l_s\big(z^{s}_{j}, q^{s}_{i}\big) =  -\sum_k q^{s, (k)}_{i} \text{ log } \dfrac{ \exp (z^{s}_{i} \cdot c^{s}_{k} / \tau)}{\sum_{k'} \exp (z^{s}_{i} \cdot c^{s}_{k'} / \tau)} \ ,
\end{gathered}
\end{equation}
where $\tau$ is a temperature hyperparameter. The objective is to maximize the agreement of prototype assignments from multiple views of one sample (RGB or flow). Features are contrasted indirectly through comparing their prototype assignments. The total loss of training the encoder $f_s$ on its own stream is taken over all videos and pairs of data augmentations, minimized with respect to both $f_s$ and $C_s$. 

\textbf{Learning stream prototype assignments.} The assignments $\{q^{s}_{i}, q^{s}_{j}\}$ are computed by matching features $\{z^{s}_{i}, z^{s}_{j}\}$ to prototypes $C_{s}$. In essence, we need to consider the cross-entropy for assigning each $z$ to $C_{s}$ and perform a mapping to assign labels automatically. Optimizing $q$ directly leads to degeneracy. Following \cite{asano_self-labelling_2020, caron_unsupervised_2020} a uniform split of the features across prototypes is enforced, which avoids the collapse of assignments to one prototype. Given our feature vectors $Z$ whose columns are $z_{1}, ..., z_{B}$, we map them to $C_s$ and optimize using an Optimal Transport \cite{optimal_transport} solver the mapping $Q = q_{1}, .., q_{B}$: 
\begin{equation}
\begin{gathered}
    \max_{ Q \in \mathcal{Q} } \ \text{Tr}\ (Q^T C_s^T Z) + \epsilon H(Q),
\end{gathered}
\end{equation}
where $H(Q)$ is the entropy of $Q$ which acts as a regularizer. The $\epsilon$ parameter controls the uniformity of the assignment where a low value helps to avoid collapse. Following \cite{caron_unsupervised_2020}, we restrict the transportation polytope to mini-batches: 
\begin{equation}
\begin{gathered}
   \mathcal{Q} = \{ Q \in \mathbb{R}^{K \times B} | \ Q \mathbbm{1}_B = \frac{1}{K} \mathbbm{1}_K, Q^T \mathbbm{1}_K = \frac{1}{B} \mathbbm{1}_K  \},
\end{gathered}
\end{equation}
where $\mathbbm{1}_K$ denotes a vector of all ones with dimension $K$. We preserve soft assignments $Q^{*}$ and the solution of the transportation polytope, solved efficiently using the Sinkhorn-Knopp algorithm \cite{cuturi_sinkhorn_2013} can be written as follows:
\begin{equation}
    Q^{*} = \text{Diag}(\alpha) \ \text{exp} \big( \frac{1}{\epsilon} C_s^T Z \big) \ \text{Diag}(\beta),
\end{equation}
where $\alpha$ and $\beta$ denote renormalization vectors such that $Q$ results in a probability matrix \cite{cuturi_sinkhorn_2013}. As the amount of batch features $B$ is usually smaller than the number of prototypes $K$, we increase or available features $B$ by adopting a queue mechanism that stores features from previous iterations. 

\subsection{Learning cross-stream}
\label{sec:crossstream}
We are now interested in using information from both streams for each encoder. Consider again the encoder $f_s$ and prototypes $C_s$ from the stream $s$ that is optimized in one alternation. We now add a second stream $t$ where $t \in \{1,2\}$ and $s$$\neq$$t$.  We use the encoder $f_t$ with \textit{frozen} weights and obtain samples $\{x^{t}_{i}, x^{t}_{j}\}$ and features $\{z^{t}_{i}, z^{t}_{j}\}$. By matching these features to prototypes $C_s$, the assignments $\{q^{t}_{i}$, $q^{t}_{j}\}$ are obtained. Given $f_s$,  $C_s$, and $f_t$, all initialized with prior representations learned on their own stream, the loss function for the prediction problem consist of four main parts:
\begin{equation} 
\begin{gathered}
\label{eq:cross}
    \Lagr_{s}^{\text{Cross-stream}} \big(z^{s}_{i}, z^{s}_{j}, z^{t}_{i}, z^{t}_{j} \big) = \\
    L_s \big(z^{s}_{i}, z^{s}_{j}, z^{t}_{j}, q^{t}_{i}\big) + \ L_s \big(z^{s}_{i}, z^{s}_{j}, z^{t}_{i}, q^{t}_{j}\big) \ + \\
    L_s \big(z^{s}_{j}, z^{t}_{i}, z^{t}_{j}, q^{s}_{i}\big)\ + \ L_s \big(z^{s}_{i}, z^{t}_{i}, z^{t}_{j}, q^{s}_{j}\big),
\end{gathered}
\end{equation}
where the function $L_s$ measures the fit between three features $z$ and an assignment $q$. For instance, the first of the $L_s$ terms is given by: 
\begin{equation} 
\begin{gathered}
    L_s\big(z^{s}_{i}, z^{s}_{j}, z^{t}_{j}, q^{t}_{i}\big) = l_s\big(z^{s}_{i}, q^{t}_{i}\big) + l_s\big(z^{s}_{j}, q^{t}_{i}\big) + l_s\big(z^{t}_{j}, q^{t}_{i}\big).
\end{gathered}
\end{equation}
The total loss function therefore consist of 12 terms. Each of the terms $l_s$ again represents the cross-entropy between one feature $z$ and one assignment $q$, \eg: 
\begin{equation} 
\begin{gathered}
    l_s(z^{s}_{i}, q^{t}_{i}) = -\sum_k q^{t, (k)}_{i} \text{ log }  \dfrac{ \exp (z^{s}_{i} \cdot c^{s}_{k} / \tau)}{\sum_{k'} \exp (z^{s}_{j} \cdot c^{s}_{k'} / \tau)},
\end{gathered}
\end{equation}
where we predict the assignment $q_i^t$ from stream $t$ (obtained by matching corresponding feature $z_i^t$ to the prototypes $C_s$) using one of the augmented features $z_i^s$ from stream $s$. 

In summary, we predict assignments from each of the four views using features originating from three views, see Figure \ref{fig:method}. In the prediction of each $q$, we avoid the use of the feature $z$ where $s$ is equal to $t$ (same stream) \textit{and} $i$ is equal to $j$ (same augmentation). This setup forces the features to capture the same information by predicting consistent assignments from them. The total loss for cross-stream training on stream $s$ is taken over all videos and pairs of augmentations, optimized with respect to $f_s$ and $C_s$. 

\textbf{Alternation.} The optimization process from this section is then performed \textit{vice versa} on the other stream. For example, we first optimize our RGB encoder $f_1$ and the corresponding prototypes $C_1$ as our $f_s$ and $C_s$ using views from both $f_s$ (RGB) and $f_t$ (flow). Next, we optimize our flow encoder $f_2$ and prototypes $C_2$ as our $f_s$ and $C_s$, and use RGB as our $f_t$. See the appendix for detailed pseudocode.

\textbf{ViCC Algorithm.} Our complete algorithm is structured as follows. \textit{Stage 1) Single-stream.} In the first stage, the two encoders $f_{1}$ and $f_{2}$ and their prototypes $C_{1}$ and $C_{2}$ are initialized from scratch and trained using their own input stream, following Equation \ref{eq:single}. \textit{Stage 2) Cross-stream.} In the second stage, cross-stream, the two models are trained in an alternating fashion using input from both streams. In one alternation, one of the streams $s$ with encoder $f_s$ and prototypes $C_s$ is encouraged to predict mappings consistently following Equation \ref{eq:cross}, leveraging complementary information from the other stream through assigning views from $f_t$ to $C_s$. Both the prototype mappings and the alternation process in our cross-stream mechanism serve as means for transferring knowledge from motion (flow) to RGB. 

\textbf{Inference.} At the inference stage, depending on speed \vs accuracy requirements, both the RGB model $f_1$ trained with ViCC self-supervision can be used for downstream tasks as well as both RGB $f_1$ and flow $f_2$ by averaging predictions from the models. 

%% file: sections/3_experiment.tex
\section{Experiments} 

\subsection{Experimental setup} 
\label{sec:expsetup}

We use two datasets for our experiments: HMDB51 \cite{kuehne_hmdb_2011} and UCF101 \cite{soomro_ucf101_2012}. UCF101 consists of 13K videos over 101 human action classes. HMDB51 is another widely used action recognition dataset and contains around 7K videos over 51 action classes. UCF101 and HMDB51 are both divided into three train/test splits. For self-supervised training we use UCF101 training split 1 without class labels. For downstream evaluation we use UCF101 and HMDB51 and evaluate on split 1 for both datasets, following prior work \cite{han_self-supervised_2020}.

\textbf{Data preprocessing.} From the source videos at 25fps, input video clips are extracted at random time stamps. Our input video clips have a spatial resolution of $128$$\times$$128$ pixels. We use clips of 32 frames as input, without temporal downsampling for S3D. For R(2+1)D and R3D, we use input clips of 16 frames with temporal downscaling at rate 2. For optical flow, we use the widely used TV-L1 algorithm \cite{zach_duality_2007} and follow practice in \cite{carreira_quo_2017, han_self-supervised_2020} for preprocessing. This means that we truncate large vectors with more than 20 in both channels, transform the values to range [0, 255] and append a third channel of 0s. Random cropping, horizontal flipping, Gaussian blur and color jittering are used in a frame-consistent manner on RGB and flow clips following recent works \cite{chen_simple_2020, he_momentum_2020}. For temporal augmentation we take clips at different time stamps with 50\% probability. 

\textbf{Implementation and training.} As our base encoder architecture we use S3D \cite{xie_rethinking_2018}. We also test our method with the R(2+1)D-18 \cite{tran_closer_2018} architecture, following recent works \cite{asano_labelling_2020, crasto_mars_2019}, and the R3D-18 \cite{hara_can_2018} backbone. We use a 2-layer MLP projection head during self-supervised training that projects the backbone output to 128 dimensional space following SimCLR \cite{chen_simple_2020}. In line with SwaV \cite{caron_unsupervised_2020}, we employ a linear layer updated by backpropagation as the prototype implementation. The projection head and the prototype layer are removed for downstream evaluation. During self-supervised training, we use a queue that consists of 1920 features. We use $K$=$300$ as the number of prototypes. The single-stream stage consists of 300 epochs. Next, the cross-steam stage is initialized with models from the single-stream stage and is trained for two cycles. In one cross-stream cycle, we first train RGB for 100 epochs and then flow for 100 epochs, each time taking the newest models, following CoCLR \cite{han_self-supervised_2020}. We run all our experiments with 4 Titan RTX GPUs with a batch size of 48. 

\textbf{Evaluation methods.} We evaluate the quality of our learned video representation using two downstream video understanding tasks: nearest neighbour video retrieval and action recognition. In the former, retrieval is performed without any supervised finetuning. We follow common protocol \cite{misra_self-supervised_2020, buchler_improving_2018, xu_self-supervised_2019} by using videos form the test set as queries for k nearest-neighbour (kNN) retrieval in the training set. We report Recall at $k$ (R@K) where we mark the retrieval as correct if a video of the same class appears among the top kNN. In the latter downstream task, we initialize with our representation and evaluate two settings: linear probe and finetuning. For linear probe, we freeze the entire network and add a linear classifier. For finetuning, the entire network with linear layer is trained end-to-end. We report Top-1 accuracy for both settings. Data augmentation similar to the self-supervision stage is used except for Gaussian blur. At inference we follow the ten-crop procedure, where the center crop, four corners and the horizontal flipped version of these crops are obtained. The moving-window approach is used for taking clips followed by averaging the predictions. 

\subsection{Model ablations}
\label{sec:ablation}

\textbf{Impact of training stages.} In Table \ref{tab:comparison} results are shown for several stages of our method in order to evaluate the improvement that cross-stream (Stage 1) has over single-stream (Stage 2). We report action recognition and nearest-neighbour video retrieval on UCF101 split 1 and include \cite{han_self-supervised_2020} as our baseline model, as it uses the contrastive instance loss on RGB and flow with additional positive examples. Training settings are kept identical across self-supervised models. All methods, including single-stream, are trained on an equal amount of epochs (500 in total). 
Evaluated on nearest-neighbour retrieval, we observe that our RGB-1 network gains a significant performance benefit when learning and predicting from optical flow in stage 2, shown as RGB-2 (62.1\% \vs 40.0\%). Furthermore, when combining predictions from the RGB-2 model and the Flow-2 model, both trained with cross-stream, we obtain a further performance boost shown as ViCC-R+F-2 (65.1\% \vs 62.1\%). We outperform \cite{han_self-supervised_2020} on retrieval by +9.5\%, demonstrating the benefit of cross-stream prototype contrasting in ViCC. In linear probe downstream classification, our RGB-2 model again outperforms the RGB-1 one by a significant margin (72.2\% \vs 49.2\%). When end-to-end finetuned our self-supervised RGB-2 outperforms RGB-1 (84.3\% \vs 81.8\%). Further improvement is found by combining the predictions of the two streams, obtaining the result for R+F (90.5\% \vs 84.3\%). Here, our performance for R+F is on par with the RGB model from \cite{han_self-supervised_2020}. As our cross-stream phase consists of cycles in which we alternate the training of streams, we further analyse the performance progress on video retrieval across training phases in Figure \ref{fig:retr_progress}. We show the evolution from single-stream to cross-stream for both models, where cross-stream consists of two cycles in which RGB and Flow are trained alternately. It can be seen that representations for both models continue to improve after one cycle, indicating that the alternating scheme is beneficial for ViCC representations.

\begin{table}
    \begin{center}
    \resizebox{1.0\columnwidth}{!}{%
    \begin{tabular}{lccccc} \specialrule{.1em}{.05em}{.05em}
    &       &            & \multicolumn{2}{c}{Classification}          & Retrieval \\ \cmidrule(lr){4-5} \cmidrule(lr){6-6}
    & &       & Linear               & Finetune & No labels \\    
     Method         & Stage & Input    & Acc                  & Acc      & R@1       \\ \specialrule{.1em}{.05em}{.05em} \Tstrut
     ViCC-RGB-1           & 1     & RGB      & 49.2                 & 81.8     & 40.0      \\
     ViCC-Flow-1           & 1     & Flow     & 71.9                 & 87.9     & 55.5      \\ \hline  \Tstrut
     \textbf{ViCC-RGB-2}          & 2     & RGB      & \textbf{72.2}                 & \textbf{84.3 }    & \textbf{62.1}      \\ 
     ViCC-Flow-2           & 2     & Flow     & 75.5                 & 88.7     & 59.7      \\ \hline  \Tstrut
     CoCLR \cite{han_self-supervised_2020} & 2 & R+F & 72.1 & 87.3 & 55.6 \\ 
     \textbf{ViCC-R+F-2}     & 2     & R+F  & \textbf{78.0 }                & \textbf{90.5} & \textbf{65.1}   \\ \specialrule{.1em}{.05em}{.05em}
    \end{tabular}
    }
    \end{center}
    \caption{\textbf{Improvement of ViCC} from single-stream (stage 1) to cross-stream (stage 2) evaluated on action recognition and nearest-neighbour retrieval on UCF101. \cite{han_self-supervised_2020} is included as a baseline comparison. R+F denotes the result obtained by averaging predictions of RGB and flow models.}
    \label{tab:comparison}
\end{table}

\begin{table}
    \begin{center}
    \resizebox{1.0\columnwidth}{!}{%
    \begin{tabular}{lcccc} \specialrule{.1em}{.05em}{.05em} 
     & \multicolumn{2}{c}{Streams for prediction}  & \multicolumn{2}{c}{Streams for assignment} \\ \cmidrule(lr){2-3} \cmidrule(lr){4-5}
     Method      & $s+t$ & $t$ & $s+t$ & $t$  \\  \specialrule{.1em}{.05em}{.05em}  \Tstrut
     ViCC-$\text{RGB-2}$      & 84.3  &  83.8   & 84.3 & 84.1 \\  \Tstrut
     ViCC-$\text{R+F-2}$     &  90.5  &  90.2  & 90.5 & 90.0 \\ \specialrule{.1em}{.05em}{.05em}
    \end{tabular}
    }
    \end{center}
    \caption{\textbf{Ablations on streams} used as views for assignment and prediction. We report Top-1 accuracy on action recognition finetuning on UCF101.}
    \label{tab:views}
\end{table}

\begin{table}
    \begin{center}
    \resizebox{0.60\columnwidth}{!}{%
    \begin{tabular}{lccc}  \specialrule{.1em}{.05em}{.05em} 
     & \multicolumn{3}{c}{Number of prototypes} \\ \cmidrule(lr){2-4} 
     Method       & 100 & 300 & 1000   \\  \specialrule{.1em}{.05em}{.05em}  \Tstrut
     ViCC-$\text{RGB-2}$   & 83.5   & 84.3  & 83.9 \\  \Tstrut
     ViCC-$\text{R+F-2}$   &  89.2   &  90.5 & 90.0  \\ \specialrule{.1em}{.05em}{.05em}
    \end{tabular}
    }
    \end{center}
    \caption{\textbf{Impact of number of prototypes}. We report Top-1 accuracy on action recognition finetuning on UCF101.}
    \label{tab:n_prot}
\end{table}

\input{figures/plot}

\input{_tables/table_linft_full}

\textbf{Ablations on stream views.} We perform an ablation study on our model by investigating the importance of the streams used as views for both prediction and assignment. We first consider the number of features for prediction, where the normal setting is to use all other available views from streams $s$ and $t$ for prediction of each assignment $q$. We now study the setting where we use two features for prediction originating only from the other stream $t$. Table \ref{tab:views} shows results for both settings, reporting Top-1 accuracy on UCF101 action recognition using the finetuning protocol. We find that using two features results in a slightly worse performance overall, suggesting that more views are beneficial for prediction of the assignments despite originating from the same stream as the assignment. The second setting that we evaluate is only using the other stream $t$ for assignment, where we map only the two features from stream $t$ to prototypes $C_s$. Note, the prediction is performed as normal, using all other available views. Both models are again slightly underperforming compared to using all views. The results for stream views in both settings suggest that the information used from the other stream in ViCC cross-stream training is of more significance than its own stream. Indeed, we find that ViCC is robust against changes in views from its own stream as it almost performs in line with results using all views for both prediction and assignment. 

\textbf{Impact of number of prototypes.} We evaluate the impact of the number of stream prototypes $K$. Explored previously by \cite{caron_unsupervised_2020} on ImageNet \cite{deng_imagenet_2009}, they found no significant impact on performance when varying the prototypes by several orders of magnitude using a sufficiently large amount of prototypes. In Table \ref{tab:n_prot}, we show results on varying the number of prototypes to $K$=$\{100, 1000\}$. We observe a slightly worse result for both settings for the RGB model and the R+F model. As we find no significant impact on the performance, our results are in line with previous work suggesting that the soft prototype mappings used for contrasting in ViCC are not necessarily a self-labeling approach similar to other pseudo-labeling approaches \cite{asano_self-labelling_2020, asano_labelling_2020, gavrilyuk_motion-augmented_2021, caron_deep_2018, yan_clusterfit_2020}, despite the usefulness in contrasting for representation learning.

\subsection{Comparison with state-of-the-art}
\label{sec:compsota}

In this section, we compare our method with self-supervised methods on action classification and video retrieval, reporting our models from the cross-stream stage.

\textbf{Action recognition.} We compare with several self-supervised methods on action recognition in Table \ref{tab:action_linft_full}, displaying our results for two backbone architectures. We organized the methods by backbone and include settings such as resolution (Res), number of frames and number of parameters (Param) for a fairer comparison. We include several methods pretrained on larger training datasets for both visual-only and multi-modal methods. In the following, we compare with visual-only modality on the same training set, with visual-only on larger datasets, and with multi-modal approaches on end-to-end finetuning. First, we significantly outperform previous approaches pretrained on UCF101 when considering the visual modality (V). On the S3D backbone, our R+F model (obtained by averaging RGB and flow predictions) achieves a Top-1 accuracy of 90.5\% on UCF101 and a Top-1 accuracy of 62.2\% on HMDB51. Our approach outperforms the best model of Han \etal \cite{han_self-supervised_2020} by 3.2\% on UCF101 and by 3.5\% on HMDB51. We also achieve better performance than Pace Pred \cite{wang_self-supervised_2020}, which uses the S3D-G \cite{xie_rethinking_2018} backbone, on both UCF101 and HMDB51. Using the R(2+1)D backbone, we obtain a Top-1 accuracy of 82.8\% on UCF101 and a Top-1 accuracy of 52.4\% on HMDB51 for RGB. When combining RGB and Flow predictions (R+F), we obtain 88.8\% and 61.5\% on the datasets respectively. We outperform VCOP \cite{luo_exploring_2020}, VCOP \cite{xu_self-supervised_2019}, PRP \cite{yao_video_2020} by a wide margin for both our models. With the R+F model we obtain a 7.2\% increase on UCF101 and a 15.1\% increase over RTT \cite{jenni_video_2020}, underlined in the table as the second-best result. ViCC models therefore consistently outperform previous works on both backbones and evaluation datasets, where optical flow provides only an optional performance boost. Comparing against visual-only information using larger training sets, we outperform methods that use Kinetics (K-400) pretraining on HMDB51, using UCF101 pretraining, such as Pace Pred \cite{wang_self-supervised_2019} for R(2+1)D and SpeedNet \cite{benaim_speednet_2020} for S3D-G. We also perform better on HMDB51 than some multi-modal approaches that use text \cite{miech_end--end_2020} and audio \cite{asano_self-labelling_2020} for similar resolution, number of frames and backbone. Finally, comparing against methods on linear probe, we outperform CoCLR \cite{han_self-supervised_2020} on the same training dataset by a significant margin. 

\input{_tables/table_retrieval_new}

\begin{figure*}
    \begin{center}
    \resizebox{1.0\textwidth}{!}{%
    \includegraphics{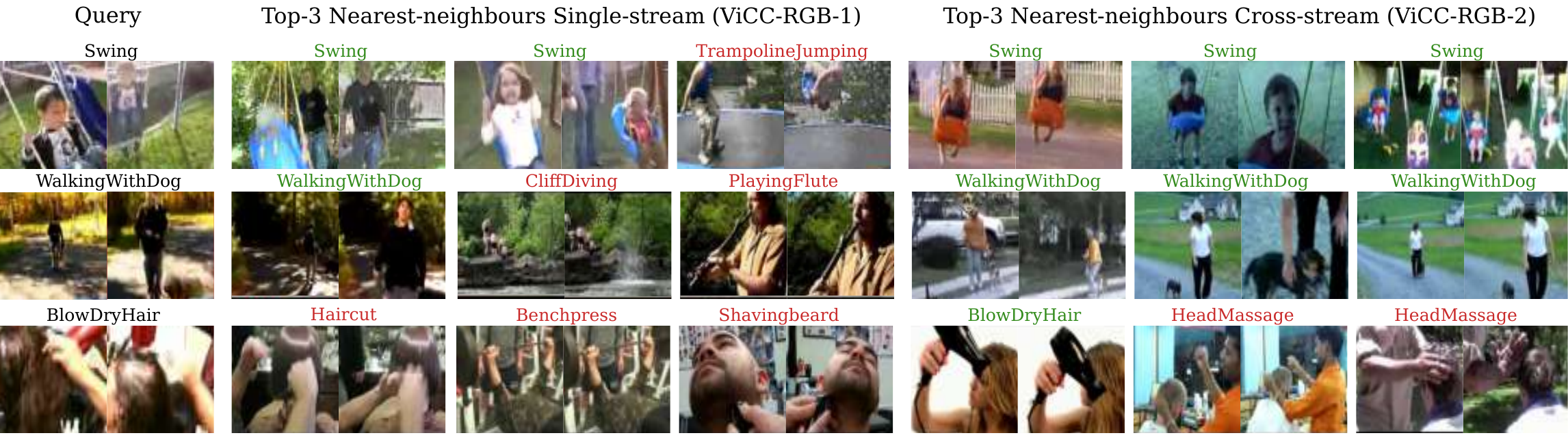}
    }
    \end{center}
    \caption{\textbf{Nearest-neighbour retrieval results with our representations}. The query video from the UCF101 test set is shown on the left, the top-3 nearest neighbours from the UCF101 training set on the right. Each video is visualized with 2 frames and we show results for single-stream (RGB-1) and cross-stream (RGB-2). The action label is shown above the video (not used during training), where {\color{correctgreen}{green}} denotes the correct label and {\color{wrongred}{red}} denotes an incorrect result. Best viewed in color.}
    \label{fig:nn}
\end{figure*}

\noindent
\textbf{Nearest-neighbour retrieval.} Next, we compare with self-supervised approaches on nearest-neighbour clip retrieval in Table \ref{tab:retrieval}. All methods are pretrained on UCF101. We also report results on R3D for a fairer comparison. Our ViCC approach outperforms all previous approaches by a significant margin on UCF101 and HMDB51 for both backbone networks R(2+1)D and S3D. Our R3D models outperforms previous methods with the same backbone significantly. We achieve a Top-1 Recall of 65.1\% on UCF101 using the S3D backbone, outperforming the previous best by 9.2\%. On HMDB51, we achieve a Top-1 Recall of 29.7\%, which is a 8.8\% increase on previous best. With the R(2+1)D backbone, we obtain a Top-1 Recall of 58.6\% on UCF101 and 25.3\% on HMDB51 for RGB, and 59.9\% and 28.3\% respectively for R+F. Compared to other self-supervised works apart from the second-best, the margins are significantly wider. We conclude that our cross-stream self-supervision model RGB learns useful motion features without needing optical flow during test time. 

\subsection{Nearest-neighbour retrieval}
\label{sec:qua}

In Figure \ref{fig:nn}, we visualize query video clips from the UCF101 test set with its Top-3 nearest-neighbours from the UCF101 training set, retrieved using the ViCC representation without labels. The ground truth action labels are included above the video clips. We visualize results for single-stream (RGB-1) and cross-stream (RGB-2). Our qualitative results further support the benefit of cross-stream training, showing that it helps to retrieve videos from the same semantic categories compared to single-stream, despite significant changes in appearance and background (\eg \textit{Swing} and \textit{WalkingWithDog}). More difficult is the retrieval for the query video from class \textit{BlowDryHair}, but we again observe that cross-stream training improves retrieval. 

%% file: figures/plot.tex
\begin{figure}
    \begin{center}
    \resizebox{0.85\columnwidth}{!}{%
    \begin{tikzpicture}[scale=1.5,
            X/.style = {circle, fill=black, inner sep=1.5pt, 
            label={[font=\scriptsize]above right:#1},
            node contents={}}
                    ]
    \begin{axis}[legend style={
                    draw=none, 
                    text depth=0pt,
                    at={(0.09,1.13)},
                    anchor=north west,
                    legend columns=-1,
                    column sep=0.5cm,
                    /tikz/column 3/.style={column sep=1pt,font=\bfseries},
                    %
                    /tikz/every odd column/.append style={column sep=0cm},
                },
                axis lines=left,
                clip=false,
                xmin = 0, xmax = 6,
                ymax=75, 
                ymin=20,
                axis y discontinuity = crunch,
                xtick distance = 1,
                xtick = {0,1,2,3,4,5,6},
                ytick = {20, 40, 60},
                yticklabels = {0, 40, 60},
                xticklabels = {{\ \ \ \ },{},{},{},{},{}},
                extra x ticks = {-4,-2},
                extra x tick style = {
                        red,
                        font=\bfseries
                        },
                grid style = {lightgray!25},
                width = \columnwidth,
                height = 0.60\columnwidth,
                grid=major, 
                grid style={dashed,gray!30}, 
                xlabel=Training phases,
                ylabel=R@1 UCF101]
    \addplot+[rgbblue,very thick,solid,mark=*,mark options={fill=rgbblue}] table [y=y, x=x, smooth, rgbblue]{figures/data/data-rgb.dat};
    \addlegendentry{RGB}
    \addplot+[floworange,very thick,solid,mark=square*,mark options={fill=floworange}] table [y=y, x=x, smooth, floworange]{figures/data/data-flow.dat};
    \addlegendentry{Flow}
    \begin{pgfonlayer}{foreground}
    \addplot+[black,very thick,dashed,mark=triangle*,mark options={fill=black}] table [y=y, x=x, smooth, black]{figures/data/data-both-1.dat};
    \addplot+[black,very thick,dashed,mark=triangle*,mark options={fill=black}] table [y=y, x=x, smooth, black]{figures/data/data-both.dat};
    \addlegendentry{R+F}
    \end{pgfonlayer}
    \addplot+[floworange,very thick,dashed,mark=square*,mark options={solid, fill=floworange}] table [y=y, x=x, smooth, floworange]{figures/data/data-flow-frozen.dat};
    \addplot+[floworange,very thick,dashed,mark=square*,mark options={solid, fill=floworange}] table [y=y, x=x, smooth, floworange]{figures/data/data-flow-frozen-1.dat};
    \addplot+[floworange,very thick,solid,mark=square*,mark options={solid, fill=floworange}] table [y=y, x=x, smooth, floworange]{figures/data/data-flow-1.dat};
    \addplot+[floworange,very thick,solid,mark=square*,mark options={solid, fill=floworange}] table [y=y, x=x, smooth, floworange]{figures/data/data-flow-2.dat};
    \addplot+[rgbblue,very thick,solid,mark=*,mark options={fill=rgbblue}] table [y=y, x=x, smooth, rgbblue]{figures/data/data-rgb-1.dat};
    \addplot+[rgbblue,very thick,solid,mark=*,mark options={fill=rgbblue}] table [y=y, x=x, smooth, rgbblue]{figures/data/data-rgb-2.dat};
    \addplot+[rgbblue,very thick,dashed,mark=*,mark options={dashed, fill=rgbblue}] table [y=y, x=x, smooth, rgbblue]{figures/data/data-rgb-frozen.dat};
    \addplot+[rgbblue,very thick,dashed,mark=*,mark options={dashed, fill=rgbblue}] table [y=y, x=x, smooth, rgbblue]{figures/data/data-rgb-frozen-2.dat};
    \node[anchor=west, color=rgbblue] at (0.95,35) {\textbf{38.0}};
    \node[anchor=west, color=rgbblue] at (1.95,65) {\textbf{60.3}};
    \node[anchor=west, color=rgbblue] at (3.95,67) {\textbf{62.1}};
    \node[anchor=west, color=floworange] at (0.9,59) {\textbf{53.4}};
    \node[anchor=west, color=floworange] at (2.95,54) {\textbf{58.5}};
    \node[anchor=west, color=floworange] at (4.95,56) {\textbf{59.7}};
    \node[anchor=west, color=black, mark=triangle*] at (4.95,56) { };
    \node[anchor=west, color=black,  draw=none, mark=triangle*] at (5.40,70) {\textbf{65.1}};
    \draw [decorate,decoration={brace,mirror,amplitude=5pt},xshift=-0.0cm,yshift=34pt]
    (0.05,0.05) -- (0.95,0.05) node [black,midway,yshift=-0.32cm, xshift=-0.25cm]
    {\footnotesize Single-stream};
     \draw [decorate,decoration={brace,mirror,amplitude=5pt},xshift=-0.0cm,yshift=34pt]
        (1.05,0.05) -- (2.95,0.05) node [black,midway,yshift=-0.32cm, xshift=0.1cm]
    {\footnotesize Cross-stream cycle};
     \draw [decorate,decoration={brace,mirror,amplitude=5pt},xshift=-0.0cm,yshift=34pt]
        (3.05,0.05) -- (4.95,0.05) node [black,midway,yshift=-0.32cm, xshift=0.15cm]
    {\footnotesize Cross-stream cycle};
     \draw [decorate,decoration={brace,mirror,amplitude=5pt},xshift=0.0cm,yshift=34pt]
        (5.05,0.05) -- (5.95, 0.05) node [black,midway,yshift=-0.30cm, xshift=0.25cm]
    {\footnotesize Combine};
    \end{axis}
    \end{tikzpicture}
    }
    \end{center}
    \caption{\textbf{Retrieval performance progress} on our training phases. RGB and flow are subsequently optimized in one cross-stream cycle, where a dotted line indicates no optimization. We report Top-1 Recall (R@1) on UCF101.}
    \label{fig:retr_progress}
\end{figure}

%% file: _tables/table_linft_full.tex
\begin{table*}
\begin{center}
\resizebox{1.0\textwidth}{!}{%
\begin{tabular}{lccccccccccc}  \specialrule{.1em}{.05em}{.05em} 
\multicolumn{8}{c}{Pretrain stage} & \multicolumn{2}{c}{Linear} & \multicolumn{2}{c}{Finetune} \\   \cmidrule(l{10pt}r{10pt}){1-8} \cmidrule(l{10pt}r{10pt}){9-10} \cmidrule(lr){11-12}              
Method                             			& Year                        & Dataset                      & Backbone                            & Param                        & Res                        & Frames                    & Modality 	& UCF101 & HMDB51 	& UCF101                      & HMDB51						\\ \specialrule{.1em}{.05em}{.05em}       
OPN \cite{lee_unsupervised_2017}   			& 2017                        & UCF101                         & VGG                             & 8.6M                         & 80                         & 16                        & V      &	-	 &      -    & 59.8                        & 23.8                       \Tstrut \\ 
VCOP \cite{xu_self-supervised_2019}  		& 2019                        & UCF101                         & R(2+1)D                         & 14.4M                        & 112                        & 16                        & V      &	-	 &     -     & 72.4                        & 30.9                        \\
Var. PSP \cite{cho_self-supervised_2020}  	& 2020                        & UCF101                         & R(2+1)D                         & 14.4M                        & 112                        & 16                        & V      &	-	 &     -     & 74.8                        & 36.8                        \\
Pace Pred \cite{wang_self-supervised_2020}  & 2020                        & UCF101                         & R(2+1)D                         & 14.4M                        & 112                        & 16                        & V      &	-	 &     -     & 75.9                        & 35.9                        \\
VCP \cite{luo_video_2020}   				& 2020                        & UCF101                         & R(2+1)D                         & 14.4M                        & 112                        & 16                        & V      &	-	 &      -    & 66.3                        & 32.2                        \\
PRP \cite{yao_video_2020}   				& 2020                        & UCF101                         & R(2+1)D                         & 14.4M                        & 112                        & 16                        & V      &	-	 &    -      & 72.1                        & 35.0                        \\
RTT \cite{jenni_video_2020}  				& 2020                        & UCF101                         & R(2+1)D                         & 14.4M                        & 112                        & 16                        & V       &	-	 &   -       & {\ul 81.6}                        & {\ul 46.4}                        \\
{\color[HTML]{9B9B9B} Pace Pred \cite{wang_self-supervised_2020}}   & {\color[HTML]{9B9B9B} 2020} & {\color[HTML]{9B9B9B} K-400} & {\color[HTML]{9B9B9B} R(2+1)D}  & {\color[HTML]{9B9B9B} 14.4M} & {\color[HTML]{9B9B9B} 112} & {\color[HTML]{9B9B9B} 16} & {\color[HTML]{9B9B9B} V}  &	-	 &        -   & {\color[HTML]{9B9B9B} 77.1} & {\color[HTML]{9B9B9B} 36.6} \\
{\color[HTML]{9B9B9B} XDC \cite{alwassel_self-supervised_2020}}     & {\color[HTML]{9B9B9B} 2020} & {\color[HTML]{9B9B9B} K-400} & {\color[HTML]{9B9B9B} R(2+1)D}  & {\color[HTML]{9B9B9B} 14.4M} & {\color[HTML]{9B9B9B} 224} & {\color[HTML]{9B9B9B} 32} & {\color[HTML]{9B9B9B} V+A} &	-	 &      -    & {\color[HTML]{9B9B9B} 86.8} & {\color[HTML]{9B9B9B} 52.6} \\
{\color[HTML]{9B9B9B} SeLaVi \cite{asano_labelling_2020}}      		& {\color[HTML]{9B9B9B} 2020} & {\color[HTML]{9B9B9B} VGG-sound \cite{chen_vggsound_2020}} & {\color[HTML]{9B9B9B} R(2+1)D}  & {\color[HTML]{9B9B9B} 14.4M} & {\color[HTML]{9B9B9B} 112} & {\color[HTML]{9B9B9B} 30} & {\color[HTML]{9B9B9B} V+A} &	-	 &    -      & {\color[HTML]{9B9B9B} 87.7} & {\color[HTML]{9B9B9B} 53.1} \\ 
{\color[HTML]{9B9B9B} GDT \cite{patrick_multi-modal_2020}}       									& {\color[HTML]{9B9B9B} 2020} & {\color[HTML]{9B9B9B} Audioset \cite{gemmeke_audio_2017}} & {\color[HTML]{9B9B9B} R(2+1)D}  & {\color[HTML]{9B9B9B} 14.4M} & {\color[HTML]{9B9B9B} 224} & {\color[HTML]{9B9B9B} 32} & {\color[HTML]{9B9B9B} V+A} &	-	 &     -      & {\color[HTML]{9B9B9B} 92.5} & {\color[HTML]{9B9B9B} 66.1} \\\hline \Tstrut
\textbf{ViCC-RGB} (\textit{ours})                			&                             & UCF101                         & R(2+1)D                         & 14.4M                        & 128                        & 16 
& V                        &	\textbf{74.4}	 &  \textbf{30.8}    	      	& \textbf{82.8}               &              \textbf{52.4}               \\ 
\textbf{ViCC-R+F} (\textit{ours})               &		                      & UCF101                         & R(2+1)D                         & 14.4M                        & 128                        & 16 
& V                        &	\textbf{78.3}	 &      \textbf{45.2}	       	& \textbf{88.8}               &    \textbf{61.5}                          \\ \hline \Tstrut
Pace Pred \cite{wang_self-supervised_2020} 							& 2020                   	  & UCF101                         & S3D-G                           & 9.6M                         & 224                        & 64                        & V          &	-	 &  -                 & {\ul 87.1}                  & {\ul 52.6}                  \\
CoCLR  \cite{han_self-supervised_2020}                            	& 2020                        & UCF101                         & S3D                             & 8.8M                         & 128                        & 32                        & V            &	70.2	 &  39.1               & 81.4                        & 52.1                        \\
CoCLR $\dagger$ \cite{han_self-supervised_2020} 								& 2020                        & UCF101                         & S3D                             & 8.8M                         & 128                        & 32                        & V             &	72.1	 &  40.2               & {\ul 87.3}                  & {\ul 58.7}                  \\
{\color[HTML]{9B9B9B} CoCLR $\dagger$ \cite{han_self-supervised_2020}}      	& {\color[HTML]{9B9B9B} 2020} & {\color[HTML]{9B9B9B} K-400} & {\color[HTML]{9B9B9B} S3D}      & {\color[HTML]{9B9B9B} 8.8M}  & {\color[HTML]{9B9B9B} 128} & {\color[HTML]{9B9B9B} 32} & {\color[HTML]{9B9B9B} V}   & {\color[HTML]{9B9B9B} 77.8} & {\color[HTML]{9B9B9B} 52.4} & {\color[HTML]{9B9B9B} 90.6} & {\color[HTML]{9B9B9B} 62.9} \\
{\color[HTML]{9B9B9B} SpeedNet \cite{benaim_speednet_2020}}      	& {\color[HTML]{9B9B9B} 2020} & {\color[HTML]{9B9B9B} K-400} & {\color[HTML]{9B9B9B} S3D-G}      & {\color[HTML]{9B9B9B} 8.8M}  & {\color[HTML]{9B9B9B} 128} & {\color[HTML]{9B9B9B} 32} & {\color[HTML]{9B9B9B} V}   & {\color[HTML]{9B9B9B} -} & {\color[HTML]{9B9B9B} -} & {\color[HTML]{9B9B9B} 81.1} & {\color[HTML]{9B9B9B} 48.8} \\ 
{\color[HTML]{9B9B9B} MIL-NCE \cite{miech_end--end_2020}}     		& {\color[HTML]{9B9B9B} 2020} & {\color[HTML]{9B9B9B} HTM \cite{miech_howto100m_2019}}   & {\color[HTML]{9B9B9B} S3D}      & {\color[HTML]{9B9B9B} 8.8M}  & {\color[HTML]{9B9B9B} 224} & {\color[HTML]{9B9B9B} 32} & {\color[HTML]{9B9B9B} V+T} & {\color[HTML]{9B9B9B} 82.7} & {\color[HTML]{9B9B9B} 53.1}& {\color[HTML]{9B9B9B} 91.3} & {\color[HTML]{9B9B9B} 61.0} \\
{\color[HTML]{9B9B9B} CBT \cite{sun_learning_2019}}         		& {\color[HTML]{9B9B9B} 2019} & {\color[HTML]{9B9B9B} K-600 \cite{carreira_short_2018}} & {\color[HTML]{9B9B9B} S3D}      & {\color[HTML]{9B9B9B} 8.8M}  & {\color[HTML]{9B9B9B} 112} & {\color[HTML]{9B9B9B} 16} & {\color[HTML]{9B9B9B} V+T} & {\color[HTML]{9B9B9B} 54.0} & {\color[HTML]{9B9B9B} 29.5}& {\color[HTML]{9B9B9B} 79.5} & {\color[HTML]{9B9B9B} 44.6} \\
\hline \Tstrut
\textbf{ViCC-RGB} (\textit{ours})                									&                             & UCF101                         & S3D                             & 8.8M                         & 128                        & 32                        & V            &	\textbf{72.2}	 &  \textbf{38.5}               & \textbf{84.3}               & \textbf{47.9}               \\
\textbf{ViCC-R+F} (\textit{ours})               						&                      		  & UCF101                         & S3D                             & 8.8M                         & 128                        & 32                        & V              &	\textbf{78.0}	 &  \textbf{47.9}              & \textbf{90.5}               & \textbf{62.2} \\  \specialrule{.1em}{.05em}{.05em} 
\end{tabular}
    }
    \end{center}
    \caption{\textbf{Comparison with prior self-supervised works on video action recognition} on UCF101 and HMDB51 for finetuning and linear probe. We report Top-1 accuracy and compare with self-supervision pretraining on UCF101. In {\color[HTML]{9B9B9B} grey} color we show larger pretraining datasets such as K-400 \cite{carreira_quo_2017} and multi-modal datasets (where T is text, A is audio). 
    }
    \label{tab:action_linft_full}
\end{table*}

%% file: _tables/table_retrieval_new.tex
\begin{table*}
    \begin{center}
    \resizebox{0.86\textwidth}{!}{%
    \begin{tabular}{lccccccccccc} \specialrule{.1em}{.05em}{.05em}
            &  &   & & \multicolumn{4}{c}{UCF101}    & \multicolumn{4}{c}{HMDB51} \\ \cmidrule(lr){5-8} \cmidrule(lr){9-12}
      Method      & Year & Backbone & Modality        & R@1            & R@5   & R@10  & R@20  & R@1             & R@5   & R@10  & R@20   \\ \specialrule{.1em}{.05em}{.05em}
      OPN   \cite{lee_unsupervised_2017}       & 2017    & VGG & V        & 19.9          & 28.7 & 34.0 & 40.6 & -               & -      &     - & -    \Tstrut \\
      ST Order \cite{buchler_improving_2018}    & 2018    & CaffeNet & V  & 25.7          & 36.2 & 42.2 & 49.2 &     -           &    -  &    -  &    -  \\
      ST-Puzzle \cite{kim_self-supervised_2019}  & 2019     & R3D & V    & 19.7   & 28.5 & 33.5 & 40.0 &  -              &    -  & -     &     - \\
      VCOP  \cite{xu_self-supervised_2019}      & 2019  & R3D  & V   & 14.1          & 30.3 & 40.4 & 51.1 & 7.6            & 22.9 & 34.4 & 48.8 \\
      Pace Pred \cite{wang_self-supervised_2020}   & 2020    & R3D  & V   & 23.8          & 38.1 & 46.4 & 56.6 &   -             &    -  &     - &    -  \\
      Var. PSP  \cite{cho_self-supervised_2020}  & 2020   & R3D   & V  & 24.6          & 41.9 & 51.3 & 62.7 &    -            &    -  &     - & -     \\
      RTT \cite{jenni_video_2020}  & 2020  & R3D   & V    & 26.1        & 48.5 & 59.1 & 69.6 &      -          &     - & -     &  -    \\ \hline \Tstrut
      \textbf{ViCC-RGB} (\textit{ours}) &          & R3D & V & 50.3 & 70.9 & 78.7 & 85.6 & 22.7 & 46.2 & 60.9 & 74.1  \\
      \textbf{ViCC-R+F} (\textit{ours})  &          & R3D  & V & 52.1  & 71.7 & 79.8 & 86.0 & 25.2 & 48.1 & 61.1 & 72.7   \\ \hline \Tstrut
            MemDPC  \cite{han_memory-augmented_2020}  & 2020    & R2D3D & V   & 20.2          & 40.4 & 52.4 & 64.7 & 7.7            & 25.7 & 40.6 & 57.7 \\ 
      VCP \cite{luo_video_2020}          & 2020  & R(2+1)D & V    & 19.9         & 33.7 & 42.0 & 50.5 & 6.7            & 21.3 & 32.7 & 49.2 \\
      CoCLR \cite{han_self-supervised_2020} & 2020   & S3D  & V      & 55.9          & 70.8 & 76.9 & 82.5 & 26.1           & 45.8 & 57.9 & 69.7 \\
    \hline \Tstrut
      \textbf{ViCC-RGB} (\textit{ours}) &          & R(2+1)D & V &  58.6  & 76.2 & 83.1 & 89.0 &   25.3 & 50.4 & 64.0 & 77.5 \\
      \textbf{ViCC-R+F} (\textit{ours})  &          & R(2+1)D  & V & 59.9  & 77.6 & 84.6 & 90.6 &   28.3 & 52.7 & 65.3 & 77.0  \\ \hline \Tstrut
            \textbf{ViCC-RGB} (\textit{ours}) &          & S3D & V       & 62.1          & 77.1 & 83.7 & 87.9 & 25.5  & 49.6 & 61.9 & 72.5 \\
      \textbf{ViCC-R+F} (\textit{ours})  &         & S3D   & V     & \textbf{65.1}          & \textbf{80.2} & \textbf{85.4} & \textbf{89.8} & \textbf{29.7} & \textbf{54.6} & \textbf{66.0} & \textbf{76.2} \\ \specialrule{.1em}{.05em}{.05em}
    \end{tabular}
    }
    \end{center}
    \caption{\textbf{Comparison with self-supervised methods on nearest-neighbour video retrieval}. All self-supervised methods are pretrained on UCF101 split 1. We show results on Top-k Recall (R@k) for $k$=$\{1, 5, 10, 20\}$ on UCF101 split 1 and HMDB51 split 1.}
    \label{tab:retrieval}
\end{table*}

%% file: sections/4_conclusion.tex
\section{Conclusion}
In this paper, we present the Video Cross-Stream Prototypical Contrasting (ViCC) framework for self-supervised representation learning. We demonstrate the advantages of using similar semantic groupings of RGB and flow views over methods that use instance-level contrastive learning, avoiding redundant comparisons and improving performance. By learning through predicting consistent prototype assignments from views originating from both streams, ViCC effectively transfers knowledge from the motion representation to appearance and vice versa. We demonstrate state-of-the-art performance on downstream video recognition tasks using visual-only self-supervision. \vspace{0.8em}

{\noindent \textbf{Acknowledgments.} We would like to thank Prof.\! dr.\! Cees G.M. Snoek for the helpful comments and feedback. \break
\noindent \textit{The presentation of this paper at the conference was financially supported by the Amsterdam ELLIS Unit, Qualcomm and the Master AI program of the University of Amsterdam.}}

%% file: sections/appendix.tex
\section{Example code for ViCC}

Here, we provide pseudocode in PyTorch-like style for the implementation of the cross-stream stage of ViCC-RGB. For the definition of the function \verb|sinkhorn| that describes the Sinkhorn-Knopp algorithm we refer to \cite{caron_unsupervised_2020}.
\begin{center}
\par\noindent\rule{\columnwidth}{0.5pt}
Pseudocode for ViCC-RGB-2 in PyTorch-like style 
\begin{lstlisting}[style=mystyle]
# rgb_model: encoder network for RGB
# flow_model: encoder network for flow, frozen
# temp: temperature
for rgb, flow in loader: # B samples
  # two augmented versions for two streams
  rgb_i, flow_i = aug(rgb_i, flow_i)
  rgb_j, flow_j = aug(rgb_j, flow_j)
  # get RGB and flow embeddings: 2B x D
  z_rgb = cat(rgb_model(rgb_i), rgb_model(rgb_j)) 
  z_flow = cat(flow_model(flow_i), flow_model(flow_j))
  # get similarity with prototypes C_rgb, C_rgb in D x K
  sim_rgb_i, sim_rgb_j = mm(z_rgb, C_rgb) 
  sim_flow_i, sim_flow_j = mm(z_flow, C_rgb) 
  # compute assignments 
  with torch.no_grad():
    q_rgb_i, q_rgb_j, q_flow_i, q_flow_j = 
    sinkhorn(sim_rgb_i), sinkhorn(sim_rgb_j), 
    sinkhorn(sim_flow_i), sinkhorn(sim_flow_j) 
  # convert similarity scores to probabilities 
  p_rgb_i, p_rgb_j, p_flow_i, p_flow_j = 
  softmax(sim_rgb_i / temp), softmax(sim_rgb_j / temp), 
  softmax(sim_flow_i / temp), softmax(sim_flow_j / temp)
  
  # predict cluster assignments using three other views
  l_rgb_i =       q_rgb_i * log(p_rgb_j) 
               + q_rgb_i * log(p_flow_i) 
               + q_rgb_i * log(p_flow_j) 
  l_rgb_j =         q_rgb_j * log(p_rgb_i) 
               + q_rgb_j * log(p_flow_i) 
               + q_rgb_j *  log(p_flow_j)
  l_flow_i =        q_flow_i * log(p_rgb_i) 
               + q_flow_i * log(p_rgb_j)  
               + q_flow_i * log(p_flow_j)
  l_flow_j =        q_flow_j * log(p_rgb_i) 
               + q_flow_j * log(p_rgb_j)  
               + q_flow_j * log(p_flow_i)
  # combine for total loss for rgb model
  loss = - 1/4 * (1/3 * l_rgb_i + 1/3 * l_rgb_j + 
                        1/3 * l_flow_i + 1/3 * l_flow_j) 
  # optimizer update and normalize prototypes
  loss.backward()
  update(rgb_model.params), update(C_rgb)
  with torch.no_grad():
    C_rgb = normalize(C_rgb, dim=0, p=2)
\end{lstlisting}
\end{center}

\section{Implementation Details}

\subsection{Implementation and Training} 
SGD with LARS \cite{you_large_2017} is used as the optimizer. A learning rate of $0.6$, a weight decay of $10^{-6}$ and a cosine learning rate schedule with a final learning rate of $6 \times 10^{-4}$ are chosen. The temperature $\tau$ is set to 0.1, the Sinkhorn regularization parameter $\epsilon$ is set to 0.05 and we perform 3 iterations of the Sinkhorn-Knopp algorithm. We use batch shuffle \cite{he_momentum_2020} to avoid the model exploiting local intra-batch information leakage for trivial solutions. For single-stream, the prototypes are frozen during the first 100 epochs of training. For cross-stream, the prototypes are directly updated from the start of the training.

\subsection{Queue}
To store additional features for use in the assignment to prototypes, we employ a queue in line with \cite{caron_unsupervised_2020}. With 4 GPUs and a total batch size of $48\times4$ = $192$, we adopt a queue of size 1920 to store features from the last 10 batches. The queue is introduced when the evolution of features is slowing down, \ie when the decrease of the loss function is moderate. For single-stream RGB (RGB-1) we introduce the queue at 150 epochs and for Flow-1 we introduce the queue at 200 epochs. For the cross-stream stage, we introduce the queue at 25 epochs in each alternation. 

\begin{figure*}
    \begin{center}
    \resizebox{0.95\textwidth}{!}{%
    \includegraphics{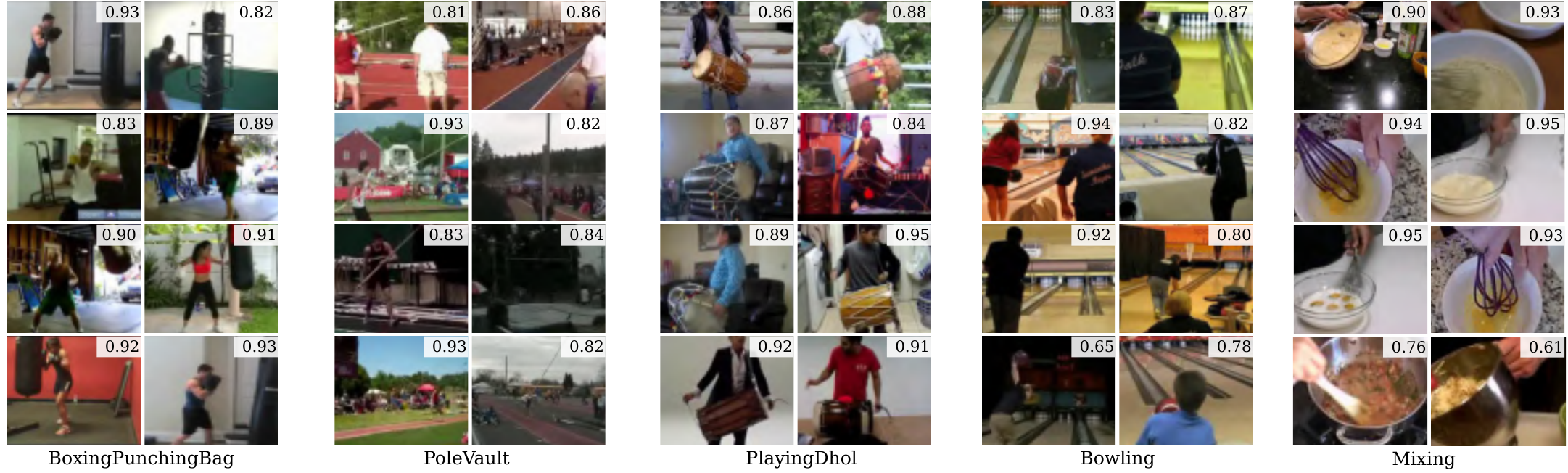}
    }
    \end{center}
    \caption{\textbf{Visualization of rounded assignments to random ViCC prototypes} using videos from UCF101. Samples with high similarity scores (visualized on the samples) to the prototypes are shown. The ground truth labels of all the video samples are included below (not used during training).}
    \label{fig:clusters}
\end{figure*}

\section{Additional results} 

\subsection{Analysis of Prototypes}

This section focuses on further analysis of the prototypes. The main purpose of the prototype sets in ViCC is to guide the contrasting of groups of views from streams in each iteration. In combination with the relatively stable performance observed when varying the number of prototypes, we conjecture that the prototypes are not a pseudo-labeling approach similar to other methods \cite{asano_self-labelling_2020, asano_labelling_2020, gavrilyuk_motion-augmented_2021, caron_deep_2018, yan_clusterfit_2020}. Despite this intuition and our use of soft assignments, we investigate the prototypes by visualizing video samples assigned to the same prototypes when rounding the assignments. We also evaluate the rounded prototype assignments from several of our self-supervised stages on standard cluster evaluation metrics. 

\subsubsection{Visualization of Prototypes} In Figure \ref{fig:clusters} we show the hard assignment of video samples to random prototypes. Video samples with the highest similarity scores to the prototype clusters are visualized. Prototype scores are indicated on the samples and the ground truth class labels of the samples are indicated below the groups. We can observe that video samples assigned to the same prototypes share semantic similarity and even belong to the same action class, despite the fact that class labels are not used during ViCC training. The prototypes seem effective at grouping together views from the same semantic class label, as the samples visualized are all from the same class. These semantically similar sets in ViCC thereby provide an advantage for video representation learning over methods that use contrastive instance learning.

\subsubsection{Cluster evaluation} 
In this section, we evaluate the hard assignment of our prototype sets with standard cluster evaluation measures as done in \cite{caron_deep_2018, asano_self-labelling_2020}. Although the ground truth number of clusters is not known in advance for self-supervised training, we set the number of prototypes to $K$=$101$ for evaluation purposes only to match the number of class labels for UCF101. The Hungarian algorithm \cite{kuhn_hungarian_1955} is then used to match self-supervised labels to the ground truth labels to obtain accuracy (Acc). We also report the Normalized Mutual Information (NMI), Adjusted Rand Index (ARI), mean entropy per cluster (where the optimal number is 0) and mean maximal purity per cluster as defined in \cite{asano_self-labelling_2020}. For example, the NMI ranges from 0 (no mutual information) to 100\% (implying perfect correlation between self-supervised labels and the ground truth labels). Table \ref{tab:clustereval} shows that our prototypes from the cross-stream stage (RGB-2 and Flow-2) obtain better performance on all measures compared to prototypes learned only on their own stream (RGB-1 and Flow-1), achieving \eg a higher NMI, lower mean entropy per cluster and higher mean maximal purity.

\begin{figure}
    \begin{center}
    \resizebox{1.0\columnwidth}{!}{%
    \includegraphics{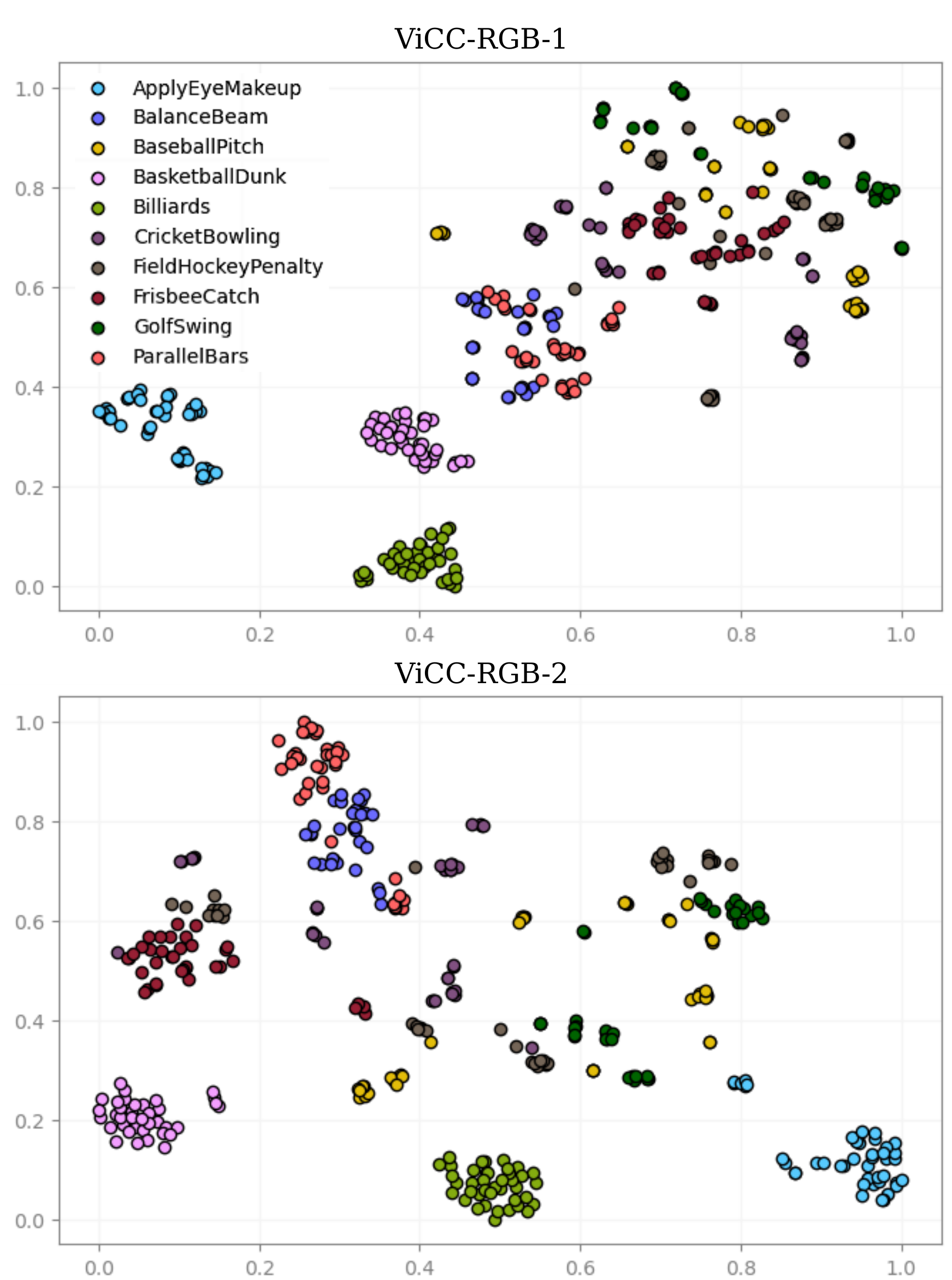}
    }
    \end{center}
    \caption{\textbf{T-SNE visualization} of the feature representations of UCF101 test set after 500 epochs of ViCC training. On the top RGB-1 single-stream is shown and on the bottom RGB-2 cross-stream.}
    \label{fig:tsne}
\end{figure}

\begin{table}[]
    \begin{center}
    \resizebox{0.95\columnwidth}{!}{%
    \begin{tabular}{lccccc} \specialrule{.1em}{.05em}{.05em}
      Method                & Acc        & NMI     & ARI         & Entropy      & Max Purity    \Tstrut   \\ \specialrule{.1em}{.05em}{.05em}
     ViCC-$\text{RGB-1}$             &  32.3      & 62.5       & 16.4          & 1.6     & 36.8 \Tstrut     \\ 
     ViCC-$\text{Flow-1}$            & 34.4       & 63.1       & 17.6          & 1.5     & 39.1      \\ 
     ViCC-$\text{RGB-2}$             & 40.8      & 67.8       &  24.5    &     1.4 &      45.1  \\  
     ViCC-$\text{Flow-2}$           & 40.3      &  67.0      & 23.5          & 1.4     &   45.3     \\ \specialrule{.1em}{.05em}{.05em}
    \end{tabular}
    }
    \end{center}
    \caption{\textbf{Cluster evaluation of ViCC prototypes} when rounding the assignments evaluated on the UCF101 test set.}
    \label{tab:clustereval}
\end{table}

\subsection{T-SNE Visualization}
In this section, we visualize ViCC representations of the UCF101 test set using the t-SNE clustering algorithm \cite{vanDerMaaten2008} to project features to 2D. For clarity, only 10 random action classes are visualized with a limited amount of random features for each class. Figure \ref{fig:tsne} shows the t-SNE visualization of features extracted from single-stream (RGB-1) and cross-stream (RGB-2) trained using the same number of epochs (500). It can be observed that the inter-class distance between certain classes such as \textit{CricketBowling} and \textit{GolfSwing} is increased from RGB-1 to RGB-2. Moreover, the intra-class distance is reduced for classes \textit{FrisbeeCatch}, \textit{BasketballDunk} and \textit{ApplyEyeMakeup}, which can be attributed to the benefit of motion learning from the flow encoder in cross-stream.   

\subsection{Impact of queue size}

We investigate the effect of the queue size on performance. The queue is used in the assignment of features to $K$ prototypes. In theory, using more features in each iteration on top of the current batch should result in a more accurate assignment for the Sinkhorn-Knopp algorithm. Results for queue sizes $\{3840, 1920, 0\}$ are shown in Table \ref{tab:queue}. We report Top-1 accuracy on action recognition on UCF101 finetuning. For queue size 3840, we observe that the larger queue size is not necessary or beneficial for UCF101 self-supervised pretraining, as the differences in performance are minimal. We also find that using no queue almost performs on par with our default queue size of 1920. We conjecture that our mini-batches may already provide enough features for ViCC self-supervision on UCF101.

\begin{table}[]
    \begin{center}
    \resizebox{0.59\columnwidth}{!}{%
    \begin{tabular}{lccc}  \specialrule{.1em}{.05em}{.05em} 
     & \multicolumn{3}{c}{Queue size} \Tstrut \\ \cmidrule(l){2-4}
     Method & 3840      & 1920  & 0  \\  \specialrule{.1em}{.05em}{.05em}  \Tstrut
     ViCC-$\text{RGB-2}$  & 84.5  &  84.3  & 84.7    \\  \Tstrut
     ViCC-$\text{R+F-2}$  & 90.4   &  90.5 & 90.2      \\ \specialrule{.1em}{.05em}{.05em}
    \end{tabular}
    }
    \end{center}
    \caption{\textbf{Impact of queue size.} We report Top-1 accuracy on action recognition finetuning on UCF101.}
    \label{tab:queue}
\end{table}

\subsection{More comparison with self-supervised works on action recognition}

In Table \ref{tab:more} we list more results from self-supervised methods evaluated on action recognition. Results for the additional backbone R3D-18 \cite{hara_can_2018} are included. We achieve better performance than several methods that use the R3D backbone. Our overall best result on the S3D backbone still outperforms almost all methods pretrained on UCF101. We also outperform several methods pretrained on the larger dataset K-400, and achieve competitive performance compared to CVRL \cite{qian_spatiotemporal_2021}.
\input{_tables/table_linft_appendix}

%% file: _tables/table_linft_appendix.tex
\begin{table*}
\begin{center}
\resizebox{1.0\textwidth}{!}{%
\begin{tabular}{lccccccccccc}  \specialrule{.1em}{.05em}{.05em} 
\multicolumn{8}{c}{Pretrain stage} & \multicolumn{2}{c}{Linear} & \multicolumn{2}{c}{Finetune} \\   \cmidrule(l{10pt}r{10pt}){1-8} \cmidrule(l{10pt}r{10pt}){9-10} \cmidrule(lr){11-12}              
Method                             			& Year                        & Dataset                      & Backbone                            & Param                        & Res                        & Frames                    & Modality 	& UCF101 & HMDB51 	& UCF101                      & HMDB51						\\ \specialrule{.1em}{.05em}{.05em}       
OPN \cite{lee_unsupervised_2017}   			& 2017                        & UCF101                         & VGG                             & 8.6M                         & 80                         & 16                        & V      &	-	 &      -    & 59.8                        & 23.8                       \Tstrut \\ 
VCOP \cite{xu_self-supervised_2019}  		& 2019                        & UCF101                         & R(2+1)D                         & 14.4M                        & 112                        & 16                        & V      &	-	 &     -     & 72.4                        & 30.9                        \\
Var. PSP \cite{cho_self-supervised_2020}  	& 2020                        & UCF101                         & R(2+1)D                         & 14.4M                        & 112                        & 16                        & V      &	-	 &     -     & 74.8                        & 36.8                        \\
Pace Pred \cite{wang_self-supervised_2020}  & 2020                        & UCF101                         & R(2+1)D                         & 14.4M                        & 112                        & 16                        & V      &	-	 &     -     & 75.9                        & 35.9                        \\
VCP \cite{luo_video_2020}   				& 2020                        & UCF101                         & R(2+1)D                         & 14.4M                        & 112                        & 16                        & V      &	-	 &      -    & 66.3                        & 32.2                        \\
PRP \cite{yao_video_2020}   				& 2020                        & UCF101                         & R(2+1)D                         & 14.4M                        & 112                        & 16                        & V      &	-	 &    -      & 72.1                        & 35.0                        \\
RTT \cite{jenni_video_2020}  				& 2020                        & UCF101                         & R(2+1)D                         & 14.4M                        & 112                        & 16                        & V       &	-	 &   -       & {\ul 81.6}                        & {\ul 46.4}                        \\
{\color[HTML]{9B9B9B} Pace Pred \cite{wang_self-supervised_2020}}   & {\color[HTML]{9B9B9B} 2020} & {\color[HTML]{9B9B9B} K-400} & {\color[HTML]{9B9B9B} R(2+1)D}  & {\color[HTML]{9B9B9B} 14.4M} & {\color[HTML]{9B9B9B} 112} & {\color[HTML]{9B9B9B} 16} & {\color[HTML]{9B9B9B} V}  &	-	 &        -   & {\color[HTML]{9B9B9B} 77.1} & {\color[HTML]{9B9B9B} 36.6} \\
{\color[HTML]{9B9B9B} MotionFit
\cite{gavrilyuk_motion-augmented_2021}}    & {\color[HTML]{9B9B9B} 2021}  & {\color[HTML]{9B9B9B} K-400} & {\color[HTML]{9B9B9B} R(2+1)D}  & {\color[HTML]{9B9B9B} 14.4M} & {\color[HTML]{9B9B9B} 112} & {\color[HTML]{9B9B9B} 32} & {\color[HTML]{9B9B9B} V}  &  {\color[HTML]{9B9B9B} -} & {\color[HTML]{9B9B9B} -} & {\color[HTML]{9B9B9B} 88.9} & {\color[HTML]{9B9B9B} 61.4} \\
{\color[HTML]{9B9B9B} XDC \cite{alwassel_self-supervised_2020}}     & {\color[HTML]{9B9B9B} 2020} & {\color[HTML]{9B9B9B} K-400} & {\color[HTML]{9B9B9B} R(2+1)D}  & {\color[HTML]{9B9B9B} 14.4M} & {\color[HTML]{9B9B9B} 224} & {\color[HTML]{9B9B9B} 32} & {\color[HTML]{9B9B9B} V+A} &	-	 &      -    & {\color[HTML]{9B9B9B} 86.8} & {\color[HTML]{9B9B9B} 52.6} \\
{\color[HTML]{9B9B9B} SeLaVi \cite{asano_labelling_2020}}      		& {\color[HTML]{9B9B9B} 2020} & {\color[HTML]{9B9B9B} VGG-sound \cite{chen_vggsound_2020}} & {\color[HTML]{9B9B9B} R(2+1)D}  & {\color[HTML]{9B9B9B} 14.4M} & {\color[HTML]{9B9B9B} 112} & {\color[HTML]{9B9B9B} 30} & {\color[HTML]{9B9B9B} V+A} &	-	 &    -      & {\color[HTML]{9B9B9B} 87.7} & {\color[HTML]{9B9B9B} 53.1} \\ 
{\color[HTML]{9B9B9B} GDT \cite{patrick_multi-modal_2020}}       									& {\color[HTML]{9B9B9B} 2020} & {\color[HTML]{9B9B9B} Audioset \cite{gemmeke_audio_2017}} & {\color[HTML]{9B9B9B} R(2+1)D}  & {\color[HTML]{9B9B9B} 14.4M} & {\color[HTML]{9B9B9B} 224} & {\color[HTML]{9B9B9B} 32} & {\color[HTML]{9B9B9B} V+A} &	-	 &     -      & {\color[HTML]{9B9B9B} 92.5} & {\color[HTML]{9B9B9B} 66.1} \\\hline \Tstrut
\textbf{ViCC-RGB} (\textit{ours})                			&                             & UCF101                         & R(2+1)D                         & 14.4M                        & 128                        & 16
& V                        &	\textbf{74.4}	 &  \textbf{30.8}    	      	& \textbf{82.8}               &              \textbf{52.4}               \\ 
\textbf{ViCC-R+F} (\textit{ours})               &		                      & UCF101                         & R(2+1)D                         & 14.4M                        & 128                        & 16
& V                        &	\textbf{78.3}	 &      \textbf{45.2}	       	& \textbf{88.8}               &    \textbf{61.5}                          \\ \hline \Tstrut
DPC  \cite{han_video_2019}                              			& 2019                        & UCF101                         & R2D3D                        & 14.2M                        & 128                        & 40                        & V                     &	-	 &     - & 60.6                        & -                           \\
MemDPC  \cite{han_memory-augmented_2020}		& 2020                        & UCF101                         & R2D3D                        & 14.2M                        & 224                        & 40                        & V                        &	-	 &    -        & 84.3 
& -                           \\
\hline \Tstrut
VCOP   \cite{xu_self-supervised_2019}     & 2019                      & UCF101                          & R3D                          & 14.2M                        & 112                        & 16                        & V                      &	-	 &    -       & 64.9                        & 29.5                        \\
Var. PSP \cite{cho_self-supervised_2020}       & 2020                  & UCF101                          & R3D                          & 14.2M                        & 112                        & 16                        & V                       &	-	 &    -      & 69.0                        & 33.7                        \\
VCP  \cite{luo_video_2020}              & 2020        & UCF101                          & R3D                          & 14.2M                        & 112                        & 16                        & V                        &	-	 &    -     & 66.0                        & 31.5                        \\
PRP \cite{yao_video_2020}                    & 2020        & UCF101                          & R3D                          & 14.2M                        & 112                        & 16                        & V                      &	-	 &    -       & 66.5                        & 29.7                        \\
RTT \cite{jenni_video_2020}          & 2020                  & UCF101                          & R3D                          & 14.2M                        & 112                        & 16                        & V                      &	-	 &    -       & {\ul 77.3}                        & {\ul 47.5}                        \\
{\color[HTML]{9B9B9B} RotNet3D \cite{jing_self-supervised_2019}}  & {\color[HTML]{9B9B9B} 2019}   & {\color[HTML]{9B9B9B} K-400} & {\color[HTML]{9B9B9B} R3D}   & {\color[HTML]{9B9B9B} 33.6M} & {\color[HTML]{9B9B9B} 224} & {\color[HTML]{9B9B9B} 16} & {\color[HTML]{9B9B9B} V} & {\color[HTML]{9B9B9B} -} & {\color[HTML]{9B9B9B} -}  & {\color[HTML]{9B9B9B} 62.9} & {\color[HTML]{9B9B9B} 33.7} \\
{\color[HTML]{9B9B9B} ST-Puzzle \cite{kim_self-supervised_2019}} & {\color[HTML]{9B9B9B} 2019}    & {\color[HTML]{9B9B9B} K-400} & {\color[HTML]{9B9B9B} R3D}   & {\color[HTML]{9B9B9B} 33.6M} & {\color[HTML]{9B9B9B} 224} & {\color[HTML]{9B9B9B} 16} & {\color[HTML]{9B9B9B} V} & {\color[HTML]{9B9B9B} -} & {\color[HTML]{9B9B9B} -}  & {\color[HTML]{9B9B9B} 65.8} & {\color[HTML]{9B9B9B} 33.7} \\
{\color[HTML]{9B9B9B} DPC \cite{han_video_2019} }     & {\color[HTML]{9B9B9B} 2019}       & {\color[HTML]{9B9B9B} K-400} & {\color[HTML]{9B9B9B} R3D}   & {\color[HTML]{9B9B9B} 14.2M} & {\color[HTML]{9B9B9B} 128} & {\color[HTML]{9B9B9B} 40} & {\color[HTML]{9B9B9B} V} & {\color[HTML]{9B9B9B} -} & {\color[HTML]{9B9B9B} -}  & {\color[HTML]{9B9B9B} 68.2} & {\color[HTML]{9B9B9B} 34.5} \\
{\color[HTML]{9B9B9B} VIE \cite{zhuang_unsupervised_2020}}  & {\color[HTML]{9B9B9B} 2020}       & {\color[HTML]{9B9B9B} K-400} & {\color[HTML]{9B9B9B} R3D}   & {\color[HTML]{9B9B9B} 14.2M} & {\color[HTML]{9B9B9B} 112} & {\color[HTML]{9B9B9B} 40} & {\color[HTML]{9B9B9B} V}  & {\color[HTML]{9B9B9B} -} & {\color[HTML]{9B9B9B} -} & {\color[HTML]{9B9B9B} 72.3} & {\color[HTML]{9B9B9B} 44.8} \\
{\color[HTML]{9B9B9B} CVRL \cite{qian_spatiotemporal_2021}} & {\color[HTML]{9B9B9B} 2021}        & {\color[HTML]{9B9B9B} K-400} & {\color[HTML]{9B9B9B} R3D-50}   & {\color[HTML]{9B9B9B} 36.1M} & {\color[HTML]{9B9B9B} 224} & {\color[HTML]{9B9B9B} 16} & {\color[HTML]{9B9B9B} V}   & {\color[HTML]{9B9B9B} -} & {\color[HTML]{9B9B9B} -} & {\color[HTML]{9B9B9B} 92.1} & {\color[HTML]{9B9B9B} 65.4} \\ \hline \Tstrut
\textbf{ViCC-RGB} (\textit{ours})                			&                             & UCF101                         & R3D                         & 14.2M                        & 128                        & 16                    & V                        &	\textbf{69.0}
& 	\textbf{44.2} 	     
& 	\textbf{78.2}       
&              	\textbf{44.7} 
\\ 
\textbf{ViCC-R+F} (\textit{ours})               &		                      & UCF101                         & R3D                         & 14.2M                        & 128                        & 16                       & V                        & \textbf{73.3} 
&     	\textbf{46.7} 	       
& \textbf{	85.7}      
&   \textbf{ 53.2} 
\\ \hline \Tstrut
Pace Pred \cite{wang_self-supervised_2020} 							& 2020                   	  & UCF101                         & S3D-G                           & 9.6M                         & 224                        & 64                        & V          &	-	 &  -                 & {\ul 87.1}                  & {\ul 52.6}                  \\
CoCLR  \cite{han_self-supervised_2020}                            	& 2020                        & UCF101                         & S3D                             & 8.8M                         & 128                        & 32                        & V            &	70.2	 &  39.1               & 81.4                        & 52.1                        \\
CoCLR $\dagger$ \cite{han_self-supervised_2020} 								& 2020                        & UCF101                         & S3D                             & 8.8M                         & 128                        & 32                        & V             &	72.1	 &  40.2               & {\ul 87.3}                  & {\ul 58.7}                  \\
{\color[HTML]{9B9B9B} CoCLR $\dagger$ \cite{han_self-supervised_2020}}      	& {\color[HTML]{9B9B9B} 2020} & {\color[HTML]{9B9B9B} K-400} & {\color[HTML]{9B9B9B} S3D}      & {\color[HTML]{9B9B9B} 8.8M}  & {\color[HTML]{9B9B9B} 128} & {\color[HTML]{9B9B9B} 32} & {\color[HTML]{9B9B9B} V}   & {\color[HTML]{9B9B9B} 77.8} & {\color[HTML]{9B9B9B} 52.4} & {\color[HTML]{9B9B9B} 90.6} & {\color[HTML]{9B9B9B} 62.9} \\
{\color[HTML]{9B9B9B} SpeedNet \cite{benaim_speednet_2020}}      	& {\color[HTML]{9B9B9B} 2020} & {\color[HTML]{9B9B9B} K-400} & {\color[HTML]{9B9B9B} S3D-G}      & {\color[HTML]{9B9B9B} 8.8M}  & {\color[HTML]{9B9B9B} 128} & {\color[HTML]{9B9B9B} 32} & {\color[HTML]{9B9B9B} V}   & {\color[HTML]{9B9B9B} -} & {\color[HTML]{9B9B9B} -} & {\color[HTML]{9B9B9B} 81.1} & {\color[HTML]{9B9B9B} 48.8} \\ 
{\color[HTML]{9B9B9B} MIL-NCE \cite{miech_end--end_2020}}     		& {\color[HTML]{9B9B9B} 2020} & {\color[HTML]{9B9B9B} HTM \cite{miech_howto100m_2019}}   & {\color[HTML]{9B9B9B} S3D}      & {\color[HTML]{9B9B9B} 8.8M}  & {\color[HTML]{9B9B9B} 224} & {\color[HTML]{9B9B9B} 32} & {\color[HTML]{9B9B9B} V+T} & {\color[HTML]{9B9B9B} 82.7} & {\color[HTML]{9B9B9B} 53.1}& {\color[HTML]{9B9B9B} 91.3} & {\color[HTML]{9B9B9B} 61.0} \\
{\color[HTML]{9B9B9B} CBT \cite{sun_learning_2019}}         		& {\color[HTML]{9B9B9B} 2019} & {\color[HTML]{9B9B9B} K-600 \cite{carreira_short_2018}} & {\color[HTML]{9B9B9B} S3D}      & {\color[HTML]{9B9B9B} 8.8M}  & {\color[HTML]{9B9B9B} 112} & {\color[HTML]{9B9B9B} 16} & {\color[HTML]{9B9B9B} V+T} & {\color[HTML]{9B9B9B} 54.0} & {\color[HTML]{9B9B9B} 29.5}& {\color[HTML]{9B9B9B} 79.5} & {\color[HTML]{9B9B9B} 44.6} \\
{\color[HTML]{9B9B9B} ELo 
\cite{piergiovanni_evolving_2020}}   & {\color[HTML]{9B9B9B} 2020}         & {\color[HTML]{9B9B9B} K-400} & {\color[HTML]{9B9B9B} S3D}      & {\color[HTML]{9B9B9B} 8.8M}  & {\color[HTML]{9B9B9B} 224} & {\color[HTML]{9B9B9B} 32} & {\color[HTML]{9B9B9B} V+T} & {\color[HTML]{9B9B9B} -} & {\color[HTML]{9B9B9B} -} & {\color[HTML]{9B9B9B} 93.8} & {\color[HTML]{9B9B9B} 67.4} \\
\hline \Tstrut
\textbf{ViCC-RGB} (\textit{ours})                									&                             & UCF101                         & S3D                             & 8.8M                         & 128                        & 32                        & V            &	\textbf{72.2}	 &  \textbf{38.5}               & \textbf{84.3}               & \textbf{47.9}               \\
\textbf{ViCC-R+F} (\textit{ours})               						&                      		  & UCF101                         & S3D                             & 8.8M                         & 128                        & 32                        & V              &	\textbf{78.0}	 &  \textbf{47.9}              & \textbf{90.5}               & \textbf{62.2} \\  \specialrule{.1em}{.05em}{.05em} 
\end{tabular}
    }
    \end{center}
    \caption{\textbf{Comparison with prior self-supervised works on video action recognition} on UCF101 and HMDB51 for finetuning and linear probe.  We report Top-1 accuracy, compare with self-supervision pretraining on UCF101 and additionally report results on backbone R3D \cite{hara_can_2018}. In {\color[HTML]{9B9B9B} grey} color we show larger pretraining datasets such as K-400 \cite{carreira_quo_2017} and multi-modal datasets (where T is text, A is audio). 
    }
    \label{tab:more}
\end{table*}